\documentclass[journal]{IEEEtran}

\ifCLASSINFOpdf
\else
\fi

\usepackage{graphicx}
\usepackage{amsmath,amssymb} % define this before the line numbering.
\usepackage{color}
\usepackage{times}
\usepackage{soul}
\usepackage{epsfig}
\usepackage{cite}
\usepackage{multirow}
\usepackage{diagbox}
\usepackage{rotating}
\usepackage{overpic}
\usepackage{subcaption}
\usepackage{booktabs}
\usepackage{array}
\usepackage{colortbl}
\usepackage[table]{xcolor}
\usepackage[linesnumbered,ruled]{algorithm2e}

\newcommand{\PreserveBackslash}[1]{\let\temp=\\#1\let\\=\temp}
\newcolumntype{C}[1]{>{\PreserveBackslash\centering}p{#1}}
\newcolumntype{R}[1]{>{\PreserveBackslash\raggedleft}p{#1}}
\newcolumntype{L}[1]{>{\PreserveBackslash\raggedright}p{#1}}

\def\eg{\emph{e.g.,~}}
\def\etal{{\em et al.~}}

\definecolor{gray1}{rgb}{.8,.8,.8}

\newcommand{\figref}[1]{Fig. \ref{#1}}
\newcommand{\tabref}[1]{Table \ref{#1}}

\newcommand{\secref}[1]{Section \ref{#1}}

\renewcommand{\arraystretch}{1.1}
\renewcommand{\tabcolsep}{.5mm}

\newcommand{\myPara}[1]{\vspace{.2in}\noindent\textbf{#1}}

\graphicspath{{./figs/}}

\usepackage[pagebackref=false,letterpaper=true,colorlinks=true,linkcolor=black,citecolor=black,bookmarks=false]{hyperref}

\begin{document}

\title{Three Birds One Stone: A General Architecture for Salient Object Segmentation, Edge Detection and Skeleton Extraction}

\author{Qibin Hou,
        Jiang-Jiang Liu,
        Ming-Ming Cheng,
        Ali Borji,
        and~Philip H.S. Torr,~\IEEEmembership{Senior~Member,~IEEE}% <-this % stops a space
%\thanks{M. Shell was with the Department
%of Electrical and Computer Engineering, Georgia Institute of Technology, Atlanta,
\thanks{Q. Hou, M.M. Cheng, and J. Liu  are with CCCE Nankai University.
M.M. Cheng is the corresponding author (cmm@nankai.edu.cn).}% <-this % stops a space
\thanks{Ali Borji is with the University of Central Florida.}
\thanks{P.H.S. Torr is with the University of Oxford.}
\thanks{Manuscript received April 19, 2005; revised August 26, 2015.}}


\markboth{Journal of \LaTeX\ Class Files,~Vol.~14, No.~8, August~2015}%
{Shell \MakeLowercase{\textit{et al.}}: Bare Demo of IEEEtran.cls for IEEE Journals}



\maketitle

\begin{abstract}

In this paper, we aim at solving pixel-wise binary problems, including
salient object segmentation, skeleton extraction, and edge detection, 
by introducing a general architecture.
Previous works have proposed tailored methods for solving each of the 
three tasks independently.
Here, we show that these tasks share some similarities that 
can be exploited for developing a general architecture.
In particular, we introduce a horizontal cascade of encoders so as to 
gradually advance the feature representations from the original CNN trunks.
To better fuse feature at different levels, 
the inputs of each encoder in our architecture
is densely connected to the outputs of its previous encoder.
Stringing these encoders together allows us to effectively exploit 
features across different levels hierarchically 
to effectively address multiple pixel-wise binary regression tasks.
To assess the performance of our proposed network on these tasks, 
we carry out exhaustive evaluations on multiple representative datasets.
Although these tasks are inherently very different, 
we show that our approach performs very well on all of them 
and works far better than current single-purpose state-of-the-art methods.
We also conduct sufficient ablation analysis to let readers 
better understand how to design encoders for different tasks.
%.
The source code in this paper will be publicly available after acceptance.

\end{abstract}


%%%%%%%%% BODY TEXT
\section{Introduction} \label{sec:introduction}

Convolutional neural networks (CNNs) have been widely used in most 
fundamental computer vision tasks (\eg
semantic segmentation~\cite{long2015fully,zhao2016pyramid}, 
edge detection \cite{xie2015holistically},
salient object segmentation~\cite{li2016deep,hou2016deeply,xu2018find,qu2017rgbd},
skeleton extraction~\cite{shen2016object,ke2017srn}.)
and have achieved unprecedented performance on many tasks.
To date, most of the existing methods are designed only for a single task
because different tasks often favor different types of features.
Their design criterion is single-purpose, 
greatly restricting their applicability to other tasks \cite{kokkinos2016ubernet}.
For example, the Holistically-nested Edge Detector (HED) \cite{xie2015holistically} 
works well for the edge detection task but does not perform well for 
salient object detection \cite{hou2016deeply}
and skeleton extraction \cite{shen2017deepskeleton}.
The reason is that the architecture proposed in HED does not consider 
how to capture homogeneous region information and scale-variant 
(either thick or thin) skeletons.

\begin{figure}
  \centering
  \includegraphics[width=\linewidth]{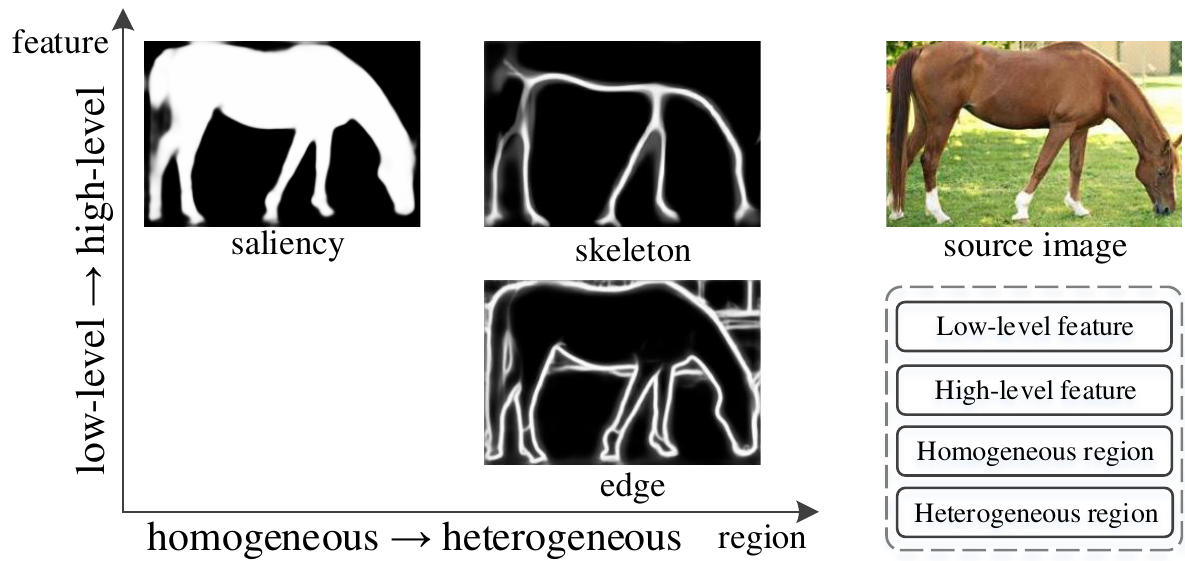}
  \caption{Preferred features of different tasks in our work. 
  	On the right side, we show the source image and two dimensions 
  	of features favored by different tasks. 
  	The results on the left side are all by our approach.
  }\label{fig:teaser}
  %\vspace{-18pt}
\end{figure}

\begin{figure*}[t]
  \centering
  \includegraphics[width=\linewidth]{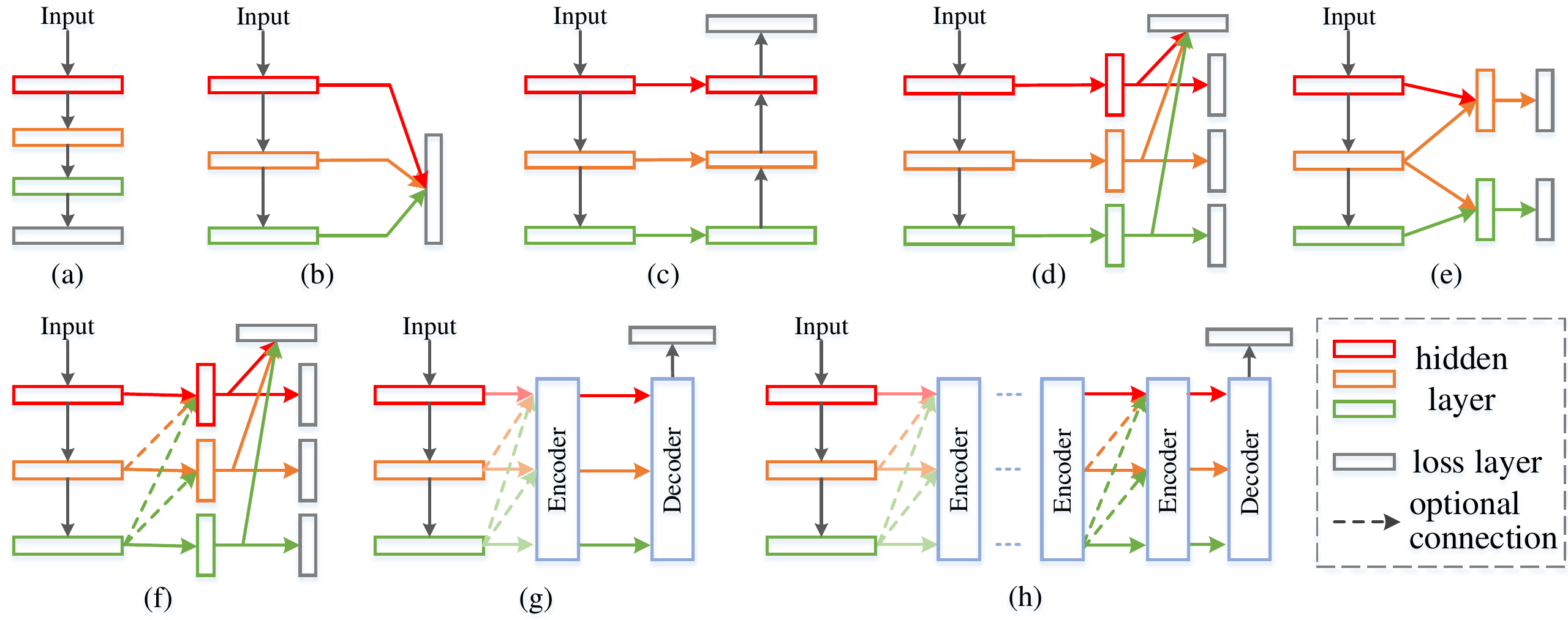}
  \caption{Architecture comparisons. (b) DCL \cite{li2016deep};
     (c) MSRNet \cite{li2017instance}; (d) HED \cite{xie2015holistically}; 
     (e) COB \cite{maninis2017convolutional}; 
     (f) SRN \cite{ke2017srn} and DSS \cite{hou2016deeply}; 
     (g) Our architecture with one encoder; 
     (h) A general case of our architecture. 
     Note that many existing methods share the same architecture 
     as one of the illustrations shown here. 
     We only show the representative one for each structure.
  }\label{fig:arch_comp}
\end{figure*}

In this paper, our goal is to present a general architecture for 
solving three important binary problems,
including salient object segmentation, edge detection, and skeleton extraction.
As popular low-level vision tasks, all of them have been widely studied recently.
In \figref{fig:teaser}, 
we illustrate a 2D space representing features that these tasks favor.
Specifically, salient object segmentation, 
as addressed in many existing works \cite{li2016deep,li2017instance,hou2016deeply,wang2018video,li2018benchmark,song2017depth},
requires the ability to extract homogeneous regions and hence relies more 
on high-level features (\figref{fig:arch_comp}c and \ref{fig:arch_comp}f).
Edge detection aims at detecting accurate boundaries, 
thus it needs more low-level features to sharpen the coarse edge maps 
produced by deeper layers \cite{xie2015holistically,maninis2017convolutional}
(\figref{fig:arch_comp}d and \ref{fig:arch_comp}e).
Skeleton extraction \cite{shen2016object,ke2017srn}, on the other hand, prefers high-level semantic information to detect scale-variant (either thick or thin) skeletons.
From the standpoint of the network architecture, in spite of three different tasks,
all of them require multi-level features in varying degrees (See \figref{fig:teaser}).
%the key difference among the aforementioned approaches
%is how many short connections (shortcut from deeper side paths to shallower ones) are needed.
%
%Edge detection and skeleton detection tend to leverage less short connections compared to
%salient object detection.
%
Consequently, a natural question is whether it is possible to combine multi-level features
in a proper way such that stronger and more general feature representations can be
constructed for solving all of these tasks.

%With regard to different features needed for segment detection and edge detection, we propose an effective way to reasonably fuse multi-level and multi-scale features
%, extracted from CNNs, to perform both tasks.
%%
%For segment detection,
%
%The following parts of this paper will be dedicated to such a challenge.

To solve the above question, 
rather than simply combining the multi-level features extracted
from the trunk of CNNs as done in most existing works
\cite{hou2016deeply,xie2015holistically,shen2017deepskeleton,dodge2018visual}, 
we propose to \emph{horizontally construct a cascade of encoders} 
to gradually encode signals from the CNN trunks (\figref{fig:arch}a)
to make the final representations more powerful.
Each encoder is composed of multiple transition nodes, 
each of which gets input from its former encoder,
enabling subsequent encoders to efficiently select features 
from the backbone in a dense manner.
As the signals from the backbone pass through the encoders sequentially, 
more and more advanced feature representations can be built 
that can be applied to different tasks.
As shown in \figref{fig:arch_comp}, 
our approach is more general compared to existing relevant methods.
To evaluate the performance of the proposed architecture, 
we apply it to three binary tasks---
salient object detection, edge detection, and skeleton extraction.
Experimental results show that our approach outperforms existing methods 
on multiple widely used benchmarks.
Specifically, for salient object detection, 
compared to previous state-of-the-art works,
our method has a performance gain of nearly 2\% on average on 5 popular datasets.
For skeleton extraction, we also significantly improve the state-of-the-art results 
by more than 2\% in terms of F-measure.
Furthermore, to let readers better understand the proposed approach, 
we conduct a series of ablation experiments for all three tasks.

To sum up, the contributions of this paper can be summarized as follows:
\begin{itemize}
  \item First, we analyse the similarities as well as differences 
  	among salient object segmentation, edge detection, and skeleton extraction 
  	and design a general architecture to solve all these tasks;
  \item Second, we propose to construct a horizontal cascade of encoders
    to progressively extract more general feature representations 
    that can be competent for all these tasks.
\end{itemize}

The rest of the paper is organized as follows.
\secref{sec:relatedWorks} reviews a number of recent works that are
strongly related to our tasks and meanwhile analyzes the differences 
among different CNN-based skip-layer architectures.
\secref{section:architecture}  presents the observations of this paper 
and describe the architecture of our proposed approach in detail.
\secref{sec:sod}-\ref{sec:skeleton} compares the proposed approach with other 
state-of-the-art results and at the same time provide sufficient ablation analysis 
to let readers better understand how each component works in our architecture. 
Finally, \secref{sec:discussion} concludes the whole paper and highlights
some potential research directions.

\section{Related Works}\label{sec:relatedWorks}

In this section, we first review a number of popular methods that are strongly related to
the tasks we solve and then analyze CNN-based skip-layer architectures
that have been proposed recently.

\subsection{Salient Object Detection}

Earlier salient object detection methods mostly rely on hand-crafted features, 
including either local contrast cues \cite{itti1998model,klein2011center,xie2013bayesian} 
or global contrast cues
\cite{cheng2015global,perazzi2012saliency,liu2011learning,jiang2013salient,huang2017300}.
Interested readers may refer to some notable review and benchmark papers~\cite{borji2014salient,borji2015salient,fan2018salient} for detailed descriptions.
Apart from the classic methods, a number of deep learning based methods have recently emerged.
In \cite{SuperCNN_IJCV2015}, He \etal presented a superpixel-wise
convolutional neural network architectures to extract hierarchical contrast features
for predicting the saliency value of each region.
Li \etal \cite{li2016visual} proposed to feed different levels of image segmentation
into three CNN branches and aggregate the multi-scale output features
to predict whether each region is salient.
Wang \etal \cite{wang2015deep} considered both local and global information
by designing two different networks.
The first one is used to learn local patch features to provide each
pixel a saliency value and then, multiple kinds of information are merged
together as the input to the second network to predict the
saliency score of each region.
Similarly, in \cite{zhao2015saliency}, Zhao \etal designed two different CNNs
to independently capture the global and local context information of each segment patch,
and then fed them into a regressor to output a saliency score for each patch.

The above methods took as input image patches and then used CNN features to predict the
saliency of each input region (either a bounding box \cite{girshick2014rich} or
a superpixel \cite{achanta2012slic,zhao2016flic}).
Later works, benefiting from the high efficiency of 
fully convolutional networks \cite{long2015fully} (FCNs),
utilize the strategy in which spatial information is processed in CNNs, 
and hence produce remarkable results.
Lee \etal \cite{lee2016deep} combined both high-level semantic features
extracted from CNNs and hand-crafted features and then utilized a unified fully-connected
neural network to estimate saliency score of each query region.
Liu \etal \cite{liu2016dhsnet,wangsaliency} refine the details of the prediction maps progressively
by harnessing recurrent fully convolutional networks.
In \cite{li2016deep}, Li \etal combined a pixel-level fully convolutional stream and
a segment-wise spatial pooling stream into one network to better leverage the
contrast information of the input images.
Hou \etal \cite{hou2016deeply} and Li \etal \cite{li2017instance} hierarchically fused
multi-level features in a top-down manner.
More recently, Zhang \etal \cite{zhang2017amulet} designed a generic architecture to
aggregating multi-level CNN features.
In \cite{zhang2017learning}, Zhang \etal embedded R-Dropout and hybrid upsampling layers
into an encoder-decoder structure to better localize the salient objects.

\subsection{Edge Detection.}

Early edge detection works \cite{canny1986computational,marr1980theory,torre1986edge} 
mostly relied on various gradient operators.
Later works, such as \cite{konishi2003statistical,martin2004learning,arbelaez2011contour}, 
were driven by manually-designed features and were able to 
improve the performance compared to gradient-based works.
Recently, with the emergence of large scale datasets, learning-based methods \cite{dollar2006supervised,ren2008multi,lim2013sketch,dollar2015fast} 
gradually became the main stream for edge detection.
Further, recent CNN-based methods
\cite{ganin2014n,shen2015deepcontour,bertasius2015deepedge,hwang2015pixel,xie2015holistically,kokkinos2015pushing,maninis2017convolutional,liu2016richer} 
have started a new wave in edge detection research.
Ganin \etal \cite{ganin2014n} proposed to use both CNN features and 
the nearest neighbour search to detect edges.
In \cite{shen2015deepcontour}, the contour data was separated into different subclasses 
and different model parameters were learned for each subclass.
Hwang \cite{hwang2015pixel} viewed edge detection as a per-pixel classification problem by learning
a feature vector for each pixel.

Different from pixel/patch level analysis methods,
Xie and Tu \cite{xie2015holistically} designed a holistically-nested edge detector (HED) 
by introducing the concept of side supervision into a fully convolutional networks 
to merge features from different levels.
In \cite{kokkinos2015pushing}, multiple cues are considered to improve the precision of edges.
Liu \etal \cite{liu2016richer} advanced the HED architecture by extracting richer 
convolutional features from CNNs.
%
%Their capacity in learning multi-level and multi-scale features allowed 
%them to achieve unprecedented performance in this field.

\subsection{Skeleton Extraction.}

Earlier methods \cite{yu2004segmentation,jang2001pseudo,majer2004influence} mainly relied on 
gradient intensity maps of natural images to extract skeletons.
Later learning-based methods viewed skeleton extraction as a per-pixel classification problem.
In \cite{tsogkas2012learning}, Tsogkas and Kokkinos calculated hand-crafted multi-scale and
multi-orientation features for each pixel and utilized multiple instance learning to predict
the score of each pixel.
Sironi \etal \cite{sironi2014multiscale} attempted to learn distance to the closest skeleton
segment for each pixel.
There are also some approaches \cite{levinshtein2013multiscale,widynski2014local} computing the similarity between superpixels and combined them
by clustering or filtering schemes.

Recent skeleton detection methods \cite{shen2016object,ke2017srn}
are mainly based on the holistically-nested edge detector (HED).
In \cite{shen2016object}, Shen \textit{et al.} introduced supervision 
in different blocks by guiding the scale-associated side outputs 
toward ground-truth skeletons at different scales and then fused
multiple scale-associated side outputs in a scale-specific manner 
to localize skeleton pixels at multiple scales.
Ke \textit{et al.} \cite{ke2017srn} added multiple shortcuts 
from deeper blocks to shallower ones based on
the HED architecture such that high-level semantic information 
can be effectively transmitted to lower side outputs, yielding stronger features.

\subsection{CNN-Based Skip-Layer Architectures}

Unlike most classification tasks which adopt the classic bottom-up structures 
(\figref{fig:arch_comp}a),
region segmentation and edge detection tasks depend on how homogeneous regions 
are extracted and how edges are sharpened.
Intuitively, considering the fact that lower network layers are capable 
of capturing local details while higher layers capture high-level contextual details, 
a good solution to satisfy the above needs might be introducing 
skip-layer architectures \cite{long2015fully}.
One of the recent successful CNN-based skip-layer structures is the HED architecture
\cite{xie2015holistically}, which learns rich hierarchical features by means
of adding side supervision to each side output (\figref{fig:arch_comp}d).
This architecture treats features at different levels equally, 
and therefore allows enough edge details to be
captured through lower side outputs.
Afterwards, several follow-up works 
\cite{kokkinos2015pushing,shen2016object,maninis2017convolutional,liu2016richer}
adopted similar structures (e.g., by introducing deep supervision) 
to capture richer feature representations by fusing features at different levels.

There are also some other work modifying this structure by adding 
short connections \cite{hou2016deeply,ke2017srn}
from upper layers to lower ones to better leverage multi-level features 
for salient object detection and skeleton extraction or 
gradually refine the coarse-level features in a top-down manner 
\cite{peng2017large,lin2017refinenet}.
These approaches, however, only attempt to simply combine multi-level features.
There is still a large room for extracting richer feature representations 
for these pixel-wise binary regression problems.

\begin{figure*}[t]
  \centering
  \includegraphics[width=\linewidth]{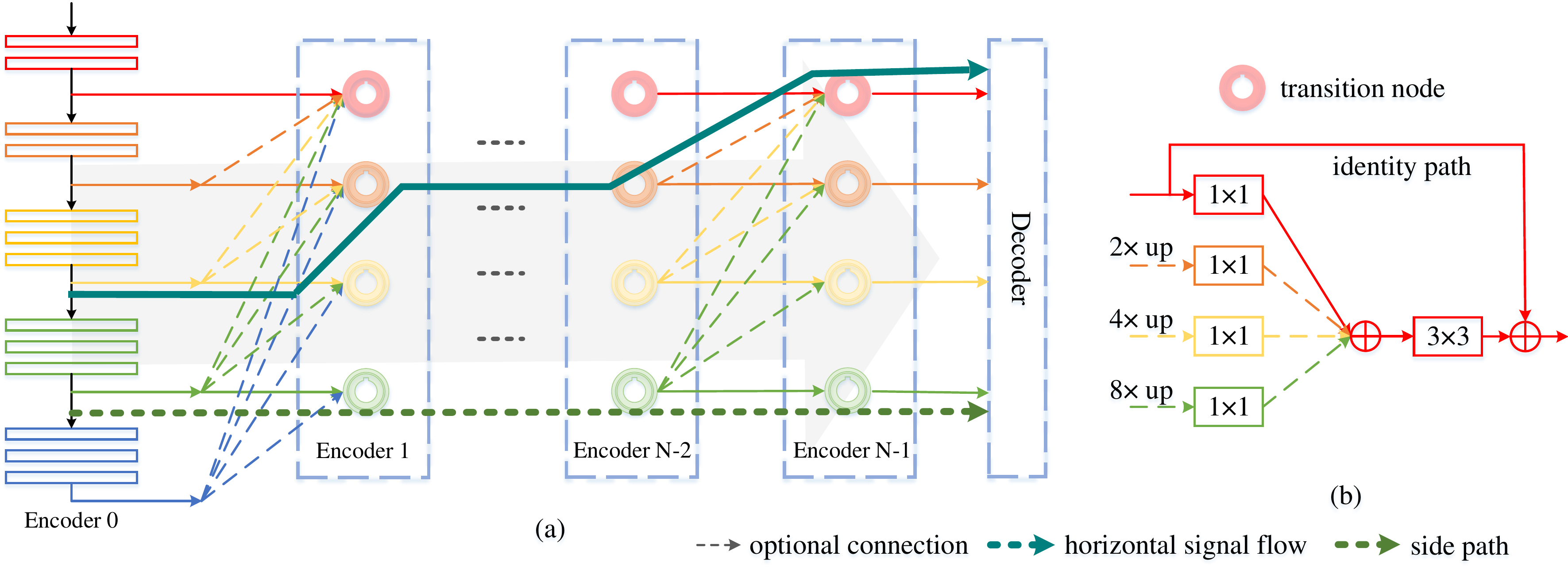}
  \caption{(a) A typical representation of our proposed general architecture. 
    (b) Detailed illustration of a transition node in each middle encoder.
    Thin solid lines are required while thin dash ones are optional. 
    Thick lines (horizontal signal flow and side path) 
    are used for demonstration purpose only.
  }\label{fig:arch}
\end{figure*}

\section{The Proposed General Architecture}
\label{section:architecture}

In this section, we elaborate on the similar characteristics 
shared by salient object detection, skeleton extraction,
and edge detection tasks and propose a general architecture that can treat them all.

\subsection{Key Observations} \label{sec:key_obs}

Previous works leverage multi-level features by either fusing simple features 
in a top-down manner (\figref{fig:arch_comp}c) or introducing shortcuts 
followed by side supervision (\figref{fig:arch_comp}d and \ref{fig:arch_comp}f).
What they have in common is that each layer in the decoder 
(the right part of each diagram in \figref{fig:arch_comp}) can only 
receive features from the backbone or its upper layers in the decoder.
These types of designs may work well for salient object detection but 
may fail when applied to edge detection and skeleton extraction (and vice versa).
The fundamental reason behind this is the fact that 
the feature representations formed in the decoders
are not powerful enough to deal with all of these tasks.

Taking into account the nature of salient object detection, 
edge detection, and skeleton extraction,
a straightforward way to build more advanced feature representations 
is to add a couple of groups of transition nodes such that multi-level features 
from the backbone can be sequentially
combined multiple times until the representations are strong enough.
In the following subsections, we will show how to construct a general structure
that can include all the features each task favors.

%The following content in this section will be dedicated to the description of our proposed framework.

%---------------------------------------------------------
\subsection{Overview of Our Proposed Architecture}
\label{sec:overview}

An illustrative diagram of our network is shown in \figref{fig:arch}a.
Structurally, our architecture can be decoupled into multiple components 
$\{S_i\}~(0\le i \le N)$, 
each of which performs different functions.
Each component can be either an encoder $\mathcal{E}$, 
encoding the feature representations from its previous
component, or a decoder $\mathcal{D}$ that decode its inputs to the final results.
Thus, the output $\mathcal{R}$ for an input $\mathcal{I}$ can be obtained by
\begin{equation}
\mathcal{R} = \mathcal{D}(\mathcal{E}_{N-1}(\cdots\mathcal{E}_1(\mathcal{E}_0(\mathcal{I}, \theta_0),\theta_1) \cdots,\theta_{N-1}), \theta_N),
\end{equation}
where $\{\theta_i\}~(0\le i \le N)$ are the learnable 
parameters for $\{S_i\}~(0\le i \le N)$, respectively.
In our architecture, the first encoder $\mathcal{E}_0$ corresponds to the backbone of 
some classification network (VGGNet \cite{simonyan2014very} here).
$\mathcal{E}_0$ is mainly used to extract the first-tier multi-level and multi-scale features
from the input images, similarly to most CNN-based architectures.
In the following sections, for convenience, we view $\mathcal{E}_1$ as our first encoder.
The decoder $\mathcal{D}$ receives signals from the last encoder $\mathcal{E}_{N-1}$ and
can have different forms depending on the task at hand.
The responsibility of the decoder is to decode the multi-level features from the 
last encoder $\mathcal{E}_{N-1}$ and output the final results.
Each middle encoder is composed of multiple transition nodes $T$, 
each of which receives input from its last
encoder and sends responses to the next component for rebuilding higher-level features.
%
%In the sequel, we say that the transition nodes in stage $S_i$ forms a group $G$.
%
For notational convenience, each block\footnote{
The definition of block here refers to the layers that share the same resolution 
in a baseline model (\eg VGGNet \cite{simonyan2014very}).} 
in $\mathcal{E}_0$ is treated as a transition node as well.
A sequence of components forms the so-called \emph{horizontal cascade}, 
which transmits the multi-level features from the backbone to the decoder horizontally.
Our architecture, obviously, can be treated as the generalized case of previous work 
(See \figref{fig:arch_comp}) and hence is quite different from them.

%---------------------------------------------------------
\subsection{Encoders}
\label{sec:trans_nodes}

%Before introducing the definition of transition node in our framework, let us first
%describe the concept of side path for better understanding.

Each middle encoder $\mathcal{E}_{i} (1\le i \le N-1)$ encodes features from its previous one
and is composed of multiple transition nodes that are used to fuse multi-level input features
in different ways.

\myPara{Side Path.} 
To better interpret the architecture of our proposed approach, 
we introduce the concept of side path in this paragraph, 
which is similar to the notation of side output in
\cite{xie2015holistically,hou2016deeply}.
Each side path, in our case, starts from the end of a block in CNNs
and ends before the decoder.
A typical example can be found in \figref{fig:arch}, 
which is represented by the dark green dash arrow with an enhanced 
thickness for highlighting.
Obviously, there are totally four standard side paths in \figref{fig:arch} 
plus a short one which is connected to the last block in $\mathcal{E}_0$.
At the beginning of each side path, we can optionally add a stack of consecutive 
convolutional layers followed by ReLU layers on each according to different tasks.
This is inspired by \cite{hou2016deeply}, who have shown that adding
more convolutional layers improves salient object detection.
In what follows, we neglect the specific number of these convolutional layers 
for the sake of convenience.
Detailed settings of each side path can be found in \secref{sec:impl_details}.

%\begin{table}
% \centering
% \begin{tabular}{C{0.5cm}C{1.2cm}C{1.8cm}C{1.5cm}C{1.0cm}C{1.8cm}} \hline
%   \# & bottom & kernel size & \#channels & stride & dilation rate \\ \hline
%   1 & conv1\_2 & $3 \times 3$ & $32$ & 1 & 1  \\
%   2 & conv2\_2 & $3 \times 3$ & $64$ & 2 & 1 \\
%   3 & conv3\_3 & $3 \times 3$ & $128$ & 4 & 1 \\
%   4 & conv4\_3 & $3 \times 3$ & $256$ & 8 & 1 \\
%   5 & conv5\_3 & $3 \times 3$ & $512$ & 16 & 1 \\
%        6 & conv6\_3 & $3 \times 3$ & $512$ & 16 & 4 \\ \hline
% \end{tabular}
% \vspace{0pt}
% \caption{Details of each side path.
% }\label{tab:side_info}
%    \vspace{-10pt}
%\end{table}

\myPara{Transition Node.}
Formally, for any positive integer $k~(k < N)$, 
let $T_m^k$ denote the $m$th transition node in $\mathcal{E}_k$.
Transition node $T_m^k$ is able to selectively receive signals (features) 
from transition nodes that are not shallower than itself 
in its previous encoder $\mathcal{E}_{k-1}$.
In this way, upper transition nodes can only get inputs from deeper side paths, 
preserving the original high-level features that are informative 
for generating homogeneous regions.
Additionally, lower transition nodes receive features from multiple levels, 
allowing these features to
be merged efficiently to produce even more advanced representations.
\figref{fig:arch}a provides an illustration,
in which transition nodes are represented by colorful solid circles, and
\figref{fig:arch}b shows a representative structure of a transition node.
%
%Notice that feature maps outputted by transition nodes on the same side path share the same level so as to
%distinguish the levels of features on different side paths.
%
Notice that, in \figref{fig:arch}a, we only show a case where the transition nodes between adjacent
encoders are densely connected.
In fact, the connection patterns can be decided according to different kinds of tasks.

\myPara{Internal Structure.}
The multi-level feature maps extracted from the backbone model 
usually contains different channel numbers and resolutions.
To ensure that features from different levels can be fused together in our architecture,
each input of a transition node is passed through a convolutional layer with the
same number of channels, followed by an upsampling layer to
make sure that all the feature maps share the same size.
The hyper-parameters of upsampling layers can be easily inferred from 
the context of our network, 
which will be elaborated in the experiments sections.
For fusion, all feature maps with the same size are merged together by simple summation.
Concatenation operation can also be used here but we empirically found that performance gain 
is negligible in all tasks.
An extra convolutional layer is added to eliminate the aliasing effect caused 
by the summation operation.
Furthermore, we also consider introducing an optional identity 
path as shown in \figref{fig:arch}b.
In this way, each transition node is allowed to automatically learn whether the signals 
from other side paths are redundant, allowing our architecture more flexible.

\subsection{Horizontal Hierarchy}
A number of recent works have leveraged multi-level features by introducing a series of
top-down paths based on the backbone of a classification network.
In our network, each transition node is able to optionally receive signals from
its previous encoder.
Unlike the fusing strategy in \cite{hou2016deeply} which only combines the score maps
from different side paths, our architecture allows more signals to be transmitted to
the next encoder or to the decoder.
In this way, each encoder is allowed to fuse features at
different levels from the backbone, allowing the output features to reach higher levels.
By adding a stack of encoders, as shown in
\figref{fig:arch}a, we are able to further advance the feature levels.
Therefore, when the number of encoders increases, a horizontal hierarchy is formed.
%To distinguish the feature levels from different stages with the vertical feature level from different blocks
%of the baseline model, we introduce the concept of horizontal hierarchy to
%describe the feature levels from different stage of transition nodes.

\myPara{Horizontal Signal Flow.}
Let $\mathcal{E}_k$ be the $k$th encoder.
%
%To cover the features from the baseline model, we use $S_0$ to denote the group formed
%by the output of different stages and $T_m^0$ the $m$th element in $G_0$.
%
The set of all transition nodes forms a multi-tree, a special directed acyclic graph (DAG)
whose vertices and edges are composed of all the transition nodes and the connections
between each pair of transition nodes.
With these definitions, a horizontal signal flow in the DAG starts from the end of an arbitrary
transition node in $\mathcal{E}_0$ and ends before the decoder.
Furthermore, the vertices through which it passes should not be lower than its starting transition node.
The dark green thick arrow in \figref{fig:arch}a depicts a representative horizontal signal flow.
When applied to different tasks, the edges can be selectively discarded.
In the following experiments sections, we will further elaborate on this and discuss
how to better leverage the horizontal signal flows for different types of binary vision tasks.

\begin{figure}[t]
    \begin{center}
        \includegraphics[width=\linewidth]{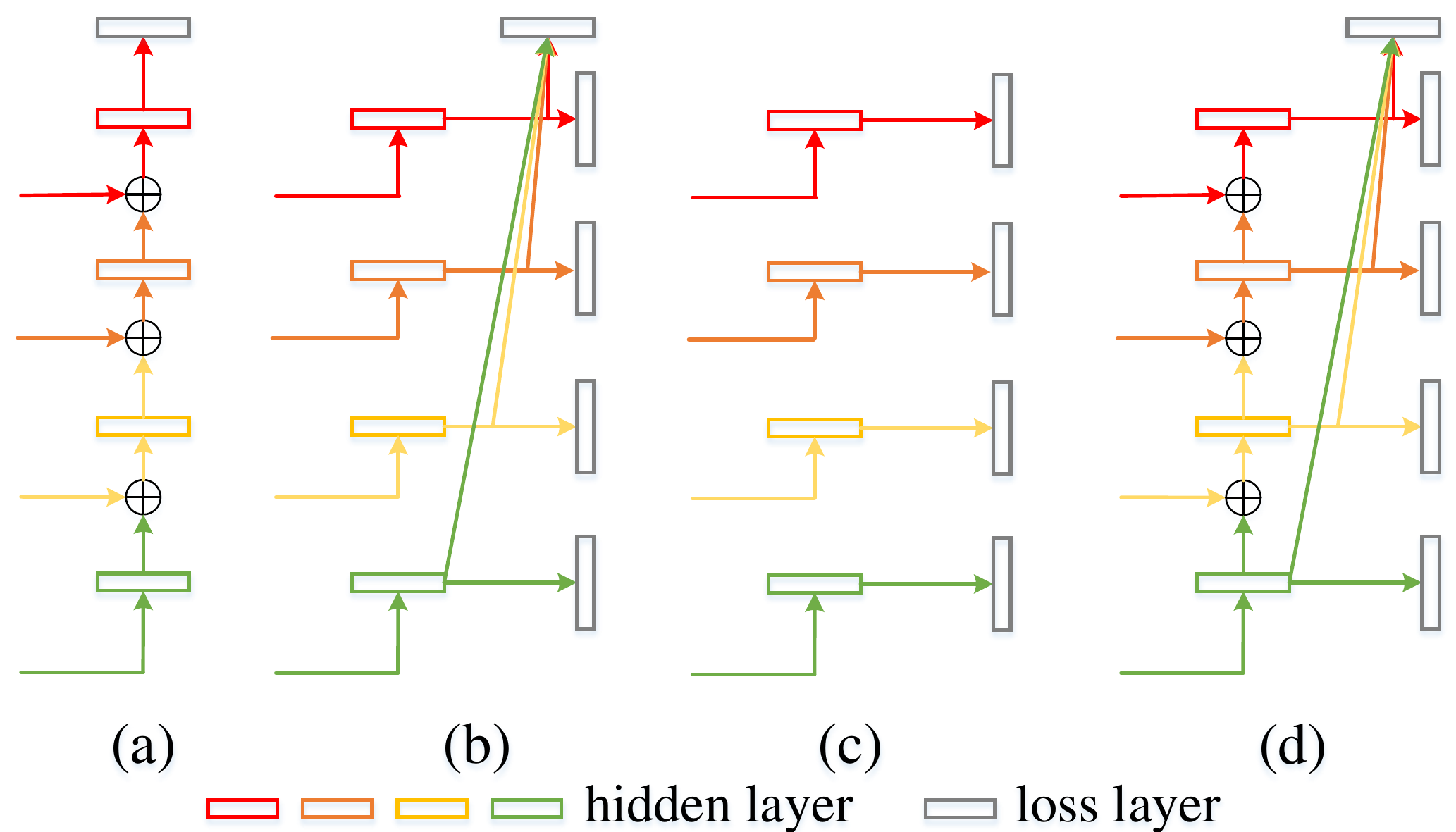}
        \caption{Different decoders.}
        \label{fig:decoders}
    \end{center}
\end{figure}

\subsection{Diverse Decoders} \label{sec:decoders}

The form of the decoder is also very important when facing different tasks.
To date, many decoders (\figref{fig:arch_comp}) with various structures have been developed.
Here, we describe two of them which we found to work very well for the three binary tasks to be solved
in this paper, respectively.
The first one corresponds to the structure in \figref{fig:arch}a, which has been adopted by many segment
detection related tasks.
This structure gradually fuses the features from different side paths in a top-down manner.
Since the feature channels from different side paths may vary, to perform the summation operation,
a convolutional layer with kernel size $3 \times 3$ is used, if needed.
In spite of only one loss layer, we found that one loss layer performs better than the structure in
\figref{fig:arch}d which introduces the concept of side supervision \cite{xie2015holistically,hou2016deeply} (See \tabref{tab:group_decoder}).
We will provide more details on the behaviors of them in the experiments sections.

For edge detection and skeleton extraction, we employ the same form of decoder as in \cite{xie2015holistically} (\figref{fig:arch}c).
Edge detection and skeleton extraction require the ability of sharpening thick lines detected and
thereby rely more on low-level features compared to segmentation.
Adding side supervision to the end of each side path allows more detailed edge information to be
emphasized.
On the other hand, the side predictions can also be merged with the weighted-fusion layer as the final output,
providing better performance.

\begin{table}[tbp!]
  \centering
  %\scriptsize
  %\footnotesize
  \renewcommand{\arraystretch}{1}
  \renewcommand{\tabcolsep}{1.9mm}
  \caption{Detailed connection information between transition nodes in adjacent encoders. Here, $T_c^{\{1, 2, 3\}}$ means that current transition node gets inputs from $T_{c-1}^1, T_{c-1}^2, T_{c-1}^3$. For edge detection and
  skeleton extraction, all the settings are the same apart from the channels numbers in each side path.
  For detailed parameter information, please refer to Sec.~\ref{sec:impl_details}.}
  \label{tab:hh_info}
  \begin{tabular}{lllllllllllllll}
  \toprule[1pt]
    & \multicolumn{3}{c}{Saliency} & \multicolumn{3}{c}{Edge \& Skeleton} \\ %& \multicolumn{2}{c}{stride} \\
   \cmidrule(l){2-4} \cmidrule(l){5-7} %\cmidrule(l){6-7}
   bottom & $S_0$ & $S_1$ & $S_2$ & $S_0$ & $S_1$ & $S_2$  \\
  \midrule[1pt]
  conv1 & $T_0^1$ & $T_1^{\{1, 2, 3, 4, 5\}}$ & $T_2^{\{1, 2, 3\}}$ & $T_0^1$ & $T_1^{\{1, 2, 3\}}$ & $T_2^{\{1, 2, 3\}}$ \\ \midrule[0pt]
  conv2 & $T_0^2$ & $T_1^{\{2, 3, 4, 5, 6\}}$ & $T_2^{\{2, 3, 4\}}$ & $T_0^2$ & $T_1^{\{2, 3, 4\}}$ & $T_2^{\{2, 3, 4\}}$ \\ \midrule[0pt]
  conv3 & $T_0^3$ & $T_1^{\{3, 4, 5, 6\}}$    & $T_2^{\{3, 4, 5\}}$ & $T_0^3$ & $T_1^{\{3, 4, 5\}}$ & $T_2^{\{3, 4\}}$ \\ \midrule[0pt]
  conv4 & $T_0^4$ & $T_1^{\{4, 5, 6\}}$       & $T_2^{\{4, 5\}}$    & $T_0^4$ & $T_1^{\{4, 5\}}$    & - \\ \midrule[0pt]
  conv5 & $T_0^5$ & $T_1^{\{5, 6\}}$          & -                   & $T_0^5$ & -                   & - \\ \midrule[0pt]
  conv6 & $T_0^6$ & -                         & -                   & -       & -                   & - \\
  \bottomrule[1pt]
  \end{tabular}
\end{table}

\begin{table*}[htp]
    \centering
    %\footnotesize
    \renewcommand{\tabcolsep}{4.5mm}
    \caption{The performance of different horizontal signal flow patterns on salient object segmentation. 
    We use the ECSSD dataset as the test set here. Pattern 1 is our detail setting. Obviously, the decrease of horizontal signal flows
    degrades the performance of our approach, which reflects the importance of dense connections between
    each pair of adjacent encoders for salient object segmentation. Note that the results listed here
    are without any post-processing tools.}
    \label{tab:sal_ablation}
    \begin{tabular}{ccccccccc} \toprule[1pt]
        %Tasks & Ours & Other Methods \\ \hline
        & \multicolumn{2}{c}{Pattern 1} & \multicolumn{2}{c}{Pattern 2} & \multicolumn{2}{c}{Pattern 3} & \multicolumn{2}{c}{Pattern 4}\\ \cmidrule[0.5pt](l){2-3} \cmidrule[0.5pt](l){4-5}\cmidrule[0.5pt](l){6-7}\cmidrule[0.5pt](l){8-9}%\cline{2-9} %\multicolumn{2}{c}{stride} \\
        %\cmidrule(l){1-2} \cmidrule(l){3-4} \cmidrule(l){5-6}
        bottom & $S_1$ & $S_2$ & $S_1$ & $S_2$ & $S_1$ & $S_2$ & $S_1$ & $S_2$ \\ \midrule[1pt]
        conv1 & $T_1^{\{1, 2, 3,4,5\}}$ & $T_2^{\{1, 2,3\}}$ & $T_1^{\{1, 2, 3\}}$ & $T_2^{\{1,2,3\}}$ & $T_1^{\{1, 2\}}$ & $T_2^{\{1, 2\}}$ & $T_1^{\{1, 2,3,4\}}$ & $T_2^{\{1, 2,3\}}$ \\
        conv2 & $T_1^{\{2, 3, 4,5,6\}}$ & $T_2^{\{2, 3,4\}}$ & $T_1^{\{2, 3, 4\}}$ & $T_2^{\{2,3,4\}}$ & $T_1^{\{2, 3\}}$ & $T_2^{\{2, 3\}}$ & $T_1^{\{2, 3,4,5\}}$ & $T_2^{\{2, 3,4\}}$   \\
        conv3 & $T_1^{\{3, 4, 5,6\}}$ & $T_2^{\{3, 4,5\}}$ & $T_1^{\{3, 4, 5\}}$ & $T_2^{\{3,4,5\}}$ & $T_1^{\{3, 4\}}$    & $T_2^{\{3, 4\}}$ & $T_1^{\{3, 4,5,6\}}$    & $T_2^{\{3, 4,5\}}$   \\
        conv4 & $T_1^{\{4, 5,6\}}$    & $T_2^{\{4, 5\}}$  & $T_1^{\{4, 5,6\}}$    & $T_2^{\{4,5\}}$ & $T_1^{\{4, 5\}}$       & $T_2^{\{4, 5\}}$ & $T_1^{\{4, 5,6\}}$       & $T_2^{\{4, 5\}}$                  \\
        conv5 & $T_1^{\{5,6\}}$ & - & $T_1^{\{5, 6\}}$ & - & $T_1^{\{5, 6\}}$ & - & $T_1^{\{5, 6\}}$ & - \\
        conv6 & - & - & - & - & - & - & - & - \\ \midrule
        Fmeasure & \multicolumn{2}{c}{0.923} & \multicolumn{2}{c}{0.918} & \multicolumn{2}{c}{0.908} & \multicolumn{2}{c}{0.921} \\ \bottomrule[1pt]
    \end{tabular}
\end{table*}

\begin{table}[t]
  \centering
    %\scriptsize
    \caption{Ablation experiments for analyzing different numbers of encoders and decoders. We use the ECSSD dataset
    as the test set here. The best results are highlighted in \textbf{bold}. No post-processing tools are used here.
  }\label{tab:group_decoder}
  \begin{tabular}{C{0.8cm}C{2.0cm}C{1.3cm}C{1.3cm}C{1.7cm}C{1.0cm}} \toprule[1pt]
    \# & Methods & \#Encoders & Decoder & F-measure & MAE \\ \midrule[1pt]
    1 & DHSNet \cite{liu2016dhsnet} & 0 & - & 0.907 & 0.059  \\
    2 & MSRNet \cite{li2017instance} & 0 & \figref{fig:arch}a & 0.913 & 0.054 \\ \midrule
    3 & GearNet (Ours)   & 1 & \figref{fig:arch}a & 0.916 & 0.055 \\
        4 & GearNet (Ours)   & 2 & \figref{fig:arch}b & 0.918 & 0.053 \\
    5 & GearNet (Ours)   & 2 & \figref{fig:arch}a & \textbf{0.923} & \textbf{0.051} \\
        \bottomrule[1pt]
  \end{tabular}
\end{table}

\subsection{Implementation Details} \label{sec:impl_details}

%We use the popular and publicly available Caffe toolbox \cite{jia2014caffe} to implement our framework.
%
As most prior works chose VGGNet \cite{simonyan2014very} as their pre-trained model, for fair comparison, we base our model on this architecture too.

For salient object detection, we replace 3 fully-connected layers with 3 convolutional layers (conv6),
the same structure to conv5.
We change the stride of pool5 to 1 and the dilatation of conv6 to 2 for large receptive fields,
and add 2 convolutional layers at the beginning of each side path following \cite{hou2016deeply}.
From side path 1 to 6, the strides are 1, 2, 4, 8, 16, and 16, respectively, and the corresponding
channel numbers of convolutional layers are set to 32, 64, 128, 256, 512, and 512, respectively.
We adopt the architecture shown in Table~\ref{tab:hh_info} as our default setting 
and the decoder in \figref{fig:decoders}a as our default decoder in our experiments.
For edge detection, we change the stride of pool4 layer to 1 and set the dilation rate of convolutional layers
in conv5 to 2 as in \cite{liu2016richer}.
The convolutional layers in each transition node are all with 16 channels by default.
The connection patterns can be found in Table~\ref{tab:hh_info}. We adopt the same decoder as in \cite{xie2015holistically,liu2016richer} (\figref{fig:decoders}c).
For skeleton extraction, the network structure is the same as in Table~\ref{tab:hh_info}
aside from the channel numbers in side paths which correspond to 32, 64, 128, and 256 from side path 1
to 4, respectively.
Thus, the strides of side paths 1 to 4 are 1, 2, 4, and 8, respectively.
All the convolutional layers mentioned here are with kernel size 3 and stride 1.

\section{Application I: Salient Object Detection}\label{sec:sod}

In this section, we apply the proposed approach to salient object segmentation. 
Salient object segmentation is an important pixel-wise binary problem 
and has attracted a lot of attention recently. We first describe the importance of
each component in our architecture by a series of ablation experiments and then
compare our method with other state-of-the-art methods.

\begin{table*}[tp!]
  \centering
  %\scriptsize
  %\footnotesize
  \renewcommand{\arraystretch}{1.0}
  \renewcommand{\tabcolsep}{3.mm}
  \caption{Quantitative salient object segmentation results over 5 widely used datasets. The best and the second best results in each column
  are highlighted in \textcolor[rgb]{0.72,0.00,0.00}{red} and \textcolor[rgb]{0.00,0.60,0.00}{green}, respectively.
  As can be seen, our approach achieves the best results on all datasets in terms of F-measure.
  }\label{tab:results}
  \begin{tabular}{lcccccccccccc}
  \toprule[1pt]
   & \multicolumn{2}{c}{Training} & \multicolumn{2}{c}{MSRA-B \cite{liu2011learning}} & \multicolumn{2}{c}{ECSSD \cite{yan2013hierarchical}} & \multicolumn{2}{c}{HKU-IS \cite{li2015visual}} & \multicolumn{2}{c}{DUT \cite{yang2013saliency}} & \multicolumn{2}{c}{SOD \cite{martin2001database,movahedi2010design}} \\
   \cmidrule(l){2-3} \cmidrule(l){4-5} \cmidrule(l){6-7} \cmidrule(l){8-9} \cmidrule(l){10-11} \cmidrule(l){12-13}
   Model & \#Images & Dataset & MaxF & MAE & MaxF & MAE & MaxF & MAE & MaxF & MAE & MaxF & MAE \\
  \midrule[1pt]
  \textbf{GC}~\cite{cheng2015global} & - & - & 0.719 & 0.159 & 0.597 & 0.233 &0.588 & 0.211 & 0.495 & 0.218 & 0.526 & 0.284 \\
  \textbf{DRFI}~\cite{cheng2015global} & 2,500 & MB & 0.845 & 0.112 & 0.782 & 0.170 & 0.776 & 0.167 & 0.664 & 0.150 & 0.699 & 0.223 \\
  \textbf{LEGS}~\cite{wang2015deep} & 3,340 & MB + P & 0.870 & 0.081 & 0.827 & 0.118 & 0.770 & 0.118 & 0.669 & 0.133 & 0.732 & 0.195 \\
  \textbf{MC}~\cite{zhao2015saliency} & 8,000 & MK & 0.894 & 0.054 & 0.837 & 0.100 & 0.798 & 0.102 & 0.703 & 0.088 & 0.727 & 0.179 \\
  \textbf{MDF}~\cite{li2015visual} & 2,500 & MB & 0.885 & 0.066 & 0.847 & 0.106 & 0.861 & 0.076 & 0.694 & 0.092 & 0.785 & 0.155 \\
  \textbf{DCL}~\cite{li2016deep} & 2,500 & MB & 0.916 & 0.047 & 0.901 & 0.068 & 0.904 & 0.049 & 0.757 & 0.080 & 0.832 & 0.126 \\
  \textbf{RFCN}~\cite{wangsaliency} &  10,000 & MK & - & - & 0.899 & 0.091 & 0.896 & 0.073 & 0.747 & 0.095 & 0.805 & 0.161  \\
  \textbf{DHSNET}~\cite{liu2016dhsnet} & 6,000 & MK & - & - & 0.905 & 0.061 & 0.892 & 0.052 & - & - & 0.823 & 0.127\\
  \textbf{ELD}~\cite{lee2016deep} & 9,000 & MK & - & - & 0.865 & 0.098 & 0.844 & 0.071 & 0.719 & 0.091 & 0.760 & 0.154  \\
  \textbf{DISC}~\cite{chen2016disc} & 9,000 & MK & 0.905 & 0.054 & 0.809 & 0.114 & 0.785 & 0.103 & 0.660 & 0.119 & - & - \\
  \textbf{MSRNet}~\cite{li2017instance} & 5,000 & MB + H & \textcolor[rgb]{0.00,0.60,0.00}{0.930} & 0.042 & 0.913 & 0.054 & \textcolor[rgb]{0.00,0.60,0.00}{0.916} & \textcolor[rgb]{0.00,0.60,0.00}{0.039} & \textcolor[rgb]{0.00,0.60,0.00}{0.785} & 0.069 & \textcolor[rgb]{0.00,0.60,0.00}{0.847} & \textcolor[rgb]{0.72,0.00,0.00}{0.112} \\
  \textbf{DSS}~\cite{hou2016deeply} & 2,500 & MB & 0.927 & \textcolor[rgb]{0.72,0.00,0.00}{0.028} & \textcolor[rgb]{0.00,0.60,0.00}{0.915} & \textcolor[rgb]{0.00,0.60,0.00}{0.052} & 0.913 & \textcolor[rgb]{0.00,0.60,0.00}{0.039} & 0.774 & \textcolor[rgb]{0.00,0.60,0.00}{0.065} & 0.842 & 0.118 \\
  \textbf{UCF}~\cite{zhang2017learning} & 10,000 & MK & - & - & 0.844 & 0.069 & 0.823 & 0.061 & 0.621 & 0.120 & 0.800 & 0.164 \\
  \textbf{Amulet}~\cite{zhang2017amulet} & 10,000 & MK & - & - & 0.868 & 0.059 & 0.843 & 0.050 & 0.647 & 0.098 & 0.801 & 0.146 \\
  \midrule[1pt]
  \textbf{GearNet} & 2,500 & MB & 0.926 & 0.039 & 0.915 & 0.060 & 0.910 & 0.044 & 0.776 & 0.069 & 0.844 & 0.124 \\
  %$\textbf{GearNet}^\text{CRF}$ & 2,500 & MB &  &  &  &  &  &  &  &  &  &  \\
  \textbf{GearNet} & 5,000 & MB + H & 0.930 & 0.039 & 0.923 & 0.055 & 0.934 & 0.034 & 0.790 & 0.068 & 0.853 & 0.117 \\
  $\textbf{GearNet}^\text{CRF}$ & 5,000 & MB + H & \textcolor[rgb]{0.72,0.00,0.00}{0.934} & \textcolor[rgb]{0.00,0.60,0.00}{0.031} & \textcolor[rgb]{0.72,0.00,0.00}{0.930} & \textcolor[rgb]{0.72,0.00,0.00}{0.046} & \textcolor[rgb]{0.72,0.00,0.00}{0.939} & \textcolor[rgb]{0.72,0.00,0.00}{0.027} & \textcolor[rgb]{0.72,0.00,0.00}{0.801} & \textcolor[rgb]{0.72,0.00,0.00}{0.058} & \textcolor[rgb]{0.72,0.00,0.00}{0.860} & \textcolor[rgb]{0.00,0.60,0.00}{0.114} \\
  \bottomrule[1pt]
  \end{tabular}
\end{table*}

\subsection{Evaluation Measures and Datasets}

Here, we use two universally-agreed, standard, and easy-to-understand
measures \cite{borji2015salient} for evaluating the existing deep saliency models.
We first report the F-measure score, which simultaneously considers recall and precision, the overlapping area
between the subjective ground truth annotation and the resulting prediction maps.
The second measure we use is the mean absolute error (MAE) between the estimated saliency map and ground-truth annotation.
%
%Detailed formulations can be found in our supplementary materials.

We perform evaluations on 5 datasets, including MSRA-B \cite{liu2011learning},
ECSSD \cite{yan2013hierarchical}, HKU-IS \cite{li2015visual},
SOD \cite{martin2001database,movahedi2010design}, and DUT-OMRON \cite{yang2013saliency}.
For training, we first use the 2,500 training images from MSRA-B.
We also try a larger training set incorporating another 2,500 images from HKU-IS as done in \cite{li2017instance}.
Notice that all the numbers reported here are from the results
the authors have presented or the results we obtained by running their publicly available code.

\subsection{Ablation Studies}

%By default, we use the architecture listed in Table~\ref{tab:hh_info}, in which there are totally
%two groups of transition nodes.
%%
%Different transition nodes are connected densely as shown in \figref{fig:arch}(a).

The hyper-parameters are as follows: weight decay (0.0005), momentum (0.9), and
mini-batch size (10).
The initial learning rate is set to 5e-3 and is divided by 10 after 8,000 iterations.
We run the network for 12,000 iterations and choose our best model according to the performance
on the validation set \cite{liu2011learning}.
Further, we also use the fully connected CRF model \cite{krahenbuhl2012efficient} which is the same as
in \cite{hou2016deeply} as a post-processing tool for maintaining spatial coherence.

\myPara{The Number of Encoders.}
The number of encoders plays an important role in our approach.
Some prior works (e.g., \cite{li2017instance,liu2016dhsnet}) have shown good results with the decoders
directly connected to the backbone.
However, when we add the first encoder, a small improvement is achieved in terms of
F-measure score.
Quantitative results can be found in Table~\ref{tab:group_decoder}.
When we add another encoder, further improvements on the F-measure and MAE scores are obtained.
We also added additional encoders but introducing more encoders yield no
further  improvements. % in F-measure, nor in MAE.
This might be due to the fact that feature representations after two times fusion have already
reached the top level of our architecture.

\myPara{More Training Data.}
The amount of training data is essential for CNN-based methods.
Besides training on the MSRA-B dataset as done in \cite{hou2016deeply,jiang2013salient,li2016deep}, we also
attempt to add more training data as in \cite{li2017instance}.
In Table~\ref{tab:results}, we show the results using different training sets.
With another 2,500 training images from \cite{li2015visual}, our performance can be
further improved about 1\% in terms of F-measure on average. % score on average.
Therefore, we believe more high-quality training data can help. %would definitely be more helpful.

\myPara{The Effect of Horizontal Signal Flows.}
By default, for salient object detection, we use a dense way to connect each pair of transition nodes
from adjacent encoders.
To show the effect of horizontal signal flows, we attempt to simplify our network by reducing the inputs of
each transition node (i.e., the dash arrows in \figref{fig:arch}a).
In Table~\ref{tab:sal_ablation}, we list four different patterns of our proposed architecture.
%
%When we adopt a similar structure to the edge part in Table~\ref{tab:hh_info},
%the F-measure performance drops slightly (about 0.5 points).
%
As can be seen, when we gradually reduce the number of horizontal signal flows, 
the performance decreases accordingly.
This phenomenon indicates that more top-down connections between
transition nodes helps segmentation type tasks.

\myPara{The Roles of Different Decoders.}
The structure of the decoder also affects the performance of our approach.
We try two different structures (Figs.~\ref{fig:arch}b and \ref{fig:arch}d) as our decoders.
Although the structure in \figref{fig:arch}b was helpful in \cite{hou2016deeply}, we obtain no performance
gain by such a structure but a slight decrease in F-measure (See \tabref{tab:group_decoder}).
%The same holds using the structure in \figref{fig:arch}d.
%
This phenomenon reveals that introducing side supervision as in \cite{hou2016deeply} is 
not always a good strategy for salient object segmentation.
Different network architectures may favor different types of decoders.

\newcommand{\addFig}[1]{\includegraphics[width=0.107\linewidth]{samples/#1}}
\newcommand{\addFigs}[1]{\addFig{#1.jpg} & \addFig{#1.png} & \addFig{#1_dtp_crf.png} &
    \addFig{#1_crf.png} & \addFig{#1_LEGS.png} & \addFig{#1_ELD.png} & \addFig{#1_rfcn.png} &
    \addFig{#1_DCL.png} & \addFig{#1_dhsnet.png}  }

\subsection{Comparison with the State-of-the-Art}

We exhaustively compare our proposed approach with 14 existing state-of-the-art salient object detection
methods including 2 classic methods (GC \cite{cheng2015global} and DRFI \cite{jiang2013salient}) and 10 recent CNN-based methods (LEGS~\cite{wang2015deep}, MC~\cite{zhao2015saliency}, MDF~\cite{li2015visual}, DCL~\cite{li2016deep}, RFCN~\cite{wangsaliency}, DHS~\cite{liu2016dhsnet}, ELD~\cite{lee2016deep}, DISC~\cite{chen2016disc}, MSRNet~\cite{li2017instance}, DSS~\cite{hou2016deeply}, UCF~\cite{zhang2017learning}, Amulet~\cite{zhang2017amulet}).
Notice that MSRNet~\cite{li2017instance} and DSS~\cite{hou2016deeply} are two of the best models to date.
Here, our best results are shown at the bottom of Table~\ref{tab:results}. % as our default scores.

\begin{figure*}[t]
    \centering
    \renewcommand{\tabcolsep}{0.28mm}
    \begin{tabular*}{1.0\textwidth}{cccccccccc}
        \addFigs{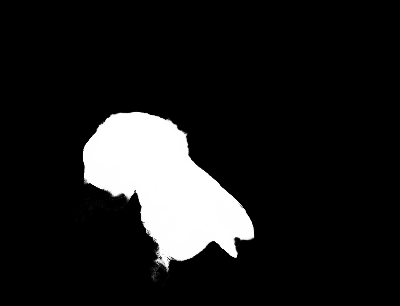} \\
        \addFigs{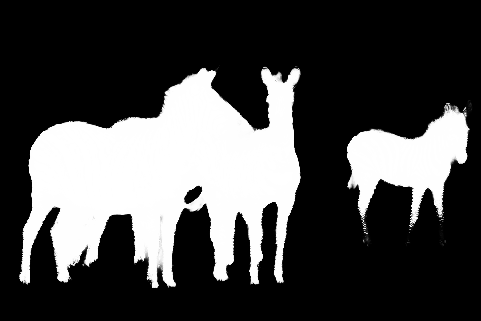} \\
        \addFigs{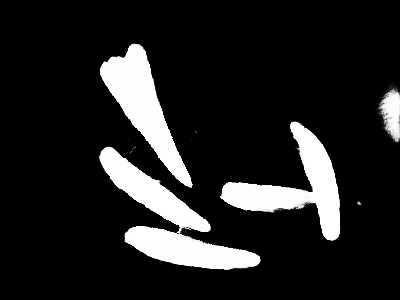} \\
        \addFigs{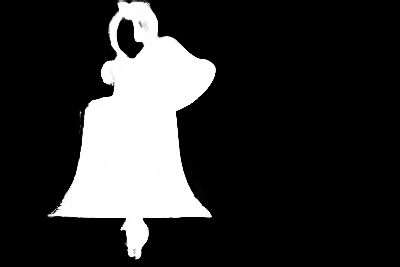} \\
        \addFigs{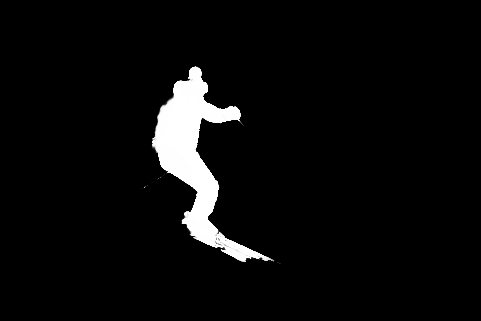} \\
        \addFigs{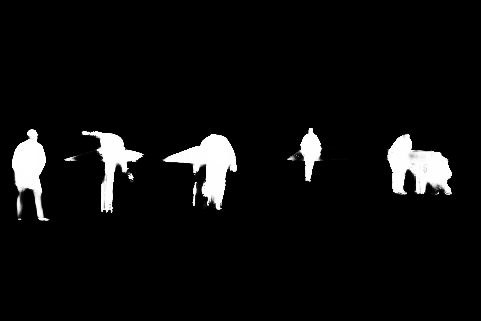} \\
        (a) Image  &  (b) GT & (c) Ours & (d) DSS & (e) LEGS & (f) ELD & (g) RFCN & (h) DCL & (i) DHS
    \end{tabular*}
    \caption{Visual comparisons of different salient object detection approaches.
    }\label{fig:visual_results}
    %\vspace{-10pt}
\end{figure*}

\myPara{F-measure and MAE Scores.}
Here, we compare our approach with the aforementioned approaches in terms of F-measure and MAE (See Table~\ref{tab:results}).
%
%In Table~\ref{tab:results}, quantitative results are shown among 12 other methods and ours.
%
As can be seen, our model trained on the MSRA-B dataset already outperforms all of the existing methods
in terms of F-measure.
With more training data, the results are improved further by a large margin (1\% on average).
This phenomenon is more pronounced when testing on the HKU-IS dataset.
Notice that our approach does better than the best existing model \cite{hou2016deeply,li2017instance}
using F-measure (2\% improvement on average).
Similar patterns can also be observed using the MAE score.

\myPara{Visual Comparisons.}
In \figref{fig:visual_results}, we show the visual comparisons with several previous state-of-the-art
approaches. In the top row, the source image have salient objects
with complex textures. As can be seen, our approach is able to successfully segment all salient objects in images
with boundaries being accurately highlighted. Some other methods such as DSS \cite{hou2016deeply} and
DCL \cite{li2016deep} also produce high quality segmentation maps, although inferior to our results. A similar phenomenon also happens when processing images where contrast between foreground and
background is low or when salient objects are tiny and irregular. Compared with existing methods, our approach performs
much better in both cases. See for example the case shown at the bottom row of \figref{fig:visual_results}. These results demonstrate that our proposed horizontal hierarchy is capable of
capturing rich and robust feature representations when applied to salient object detection.
%In addition to \figref{fig:visual_results}, we also provide more visual results which can be found in our
%supplementary materials.

%\myPara{Failure Case Analysis.}

\section{Application II: Edge Detection}\label{sec:edge}

%[[An opening would be nice here!!]]

In this section, we apply our GearNet to edge detection, 
one of the popular and basic low-level tasks in computer vision.

%\subsection{Implementation Details} \label{sec:edge_impl}
%
%Similar to most prior works \cite{xie2015holistically,liu2016richer}, we also adopt the 16-layer VGGNet
%\cite{simonyan2014very} pretrained on ImageNet \cite{krizhevsky2012imagenet} as our backbone here.
%%
%A detailed configuration of our edge model can be found in Table~\ref{tab:hh_info}.
%%
%We keep all the convolutional layers of VGGNet while drop out all the fully connected layers as in \cite{xie2015holistically}.
%%
%To preserve fine resolution of feature maps in each path, we change the stride of pool4 layer to 1
%and set the dilation rate of convolutional layers in the 5th block to 2 for receiving large receptive fields.
%%
%At the beginning of each side path, we do not add any convolutional layers as suggested by Xie et al. \cite{xie2015holistically}.
%%
%The convolutional layers in each transition node are all with 8 channels, kernel size 3 and stride 1.
%%
%About the channel numbers here, we found that using more channels has little effect on the performance but
%increases the runtime cost.
%%
%About the decoder, we adopt the same one as in \cite{xie2015holistically,liu2016richer} (\figref{fig:arch}c),
%which we found achieves the best performance compared to others in \figref{fig:arch}.

\begin{table}[!t]
    \centering
    %\scriptsize
    \renewcommand{\arraystretch}{.8}
    \caption{Quantitative comparison of our approach with existing edge detection methods. Here, `MS' means multi-scale test as in \cite{liu2016richer}. Here, we use three scales \{0.5, 1.0, 1.5, 2.0\}.
        %`Fuse' means that we combine the output from each side path and the fused output.
        `I' and `II' correspond to the networks that are with 1 encoder and 2 encoders, respectively. The best results are highlighted in \textbf{bold}.}
    \begin{tabular}{L{40mm}C{20mm}C{20mm}} \toprule[1pt]%\whline{1pt}%\cline{1-2}
      & \multicolumn{2}{c}{Edge}  \\ \cmidrule(l){2-3}
      Method & ODS & OIS \\ \midrule[1pt]
        gPb-owt-ucm \cite{arbelaez2011contour}  & 0.726 & 0.757 \\ \midrule[-0.7mm]
        SE-Var \cite{dollar2015fast}  & 0.746 & 0.767 \\ \midrule[-0.7mm]
        MCG \cite{pont2017multiscale}  & 0.747 & 0.779 \\ \midrule
        %----------------------------------------------------------------
        DeepEdge \cite{bertasius2015deepedge}  & 0.753 & 0.772 \\  \midrule[-0.7mm]
      DeepContour \cite{shen2015deepcontour}  & 0.756 & 0.773 \\ \midrule[-0.7mm]
        HED \cite{xie2015holistically}  & 0.788 & 0.808 \\ \midrule[-0.7mm]
        CEDN \cite{yang2016object}  & 0.788 & 0.804 \\ \midrule[-0.7mm]
        RDS \cite{liu2016learning} & 0.792 & 0.810 \\  \midrule[-0.7mm]
        COB \cite{maninis2017convolutional} & 0.793 & 0.820 \\  \midrule[-0.7mm]
        CED \cite{wang2017deep} & 0.803 & 0.820 \\ \midrule[-0.7mm]
        DCNN+sPb \cite{kokkinos2015pushing}  & 0.813 & 0.831 \\ \midrule[-0.7mm]
        RCF-MS \cite{liu2016richer}  & 0.811 & 0.830 \\ \midrule
        %----------------------------------------------------------------
        %$\text{GearNet}^{\mathrm{II}*}$ (Ours) & 60.2\% & 61.4\% & - \\ \midrule[-0.7mm]
        %$\text{GearNet}^{\mathrm{II}}$ (Ours) & - & - \\ \midrule[-0.7mm]
        $\text{GearNet-MS}^{\mathrm{I}}$ (Ours) & 0.815 & 0.834 \\ \midrule[-0.7mm]
        $\text{GearNet-MS}^{\mathrm{II}}$ (Ours) & \textbf{0.818} & \textbf{0.836} \\
        %$\text{GearNet-MS}^{\mathrm{II}}$ (Ours) & 0.815 & 0.835 \\ \midrule[-0.7mm]
        %$\text{GearNet-MS-Fuse}^{\mathrm{II}}$  (Ours) & \textbf{0.818} & \textbf{0.836} \\
        %$\text{GearNet-MS-Fuse}^{\mathrm{I}}$ (Ours) & 0.815 & 0.834 \\
        \bottomrule[1pt]
        \end{tabular}

        \label{tab:edge_comps}
\end{table}

The hyper-parameters used in our experiment include mini-batch size set to 10, momentum set to 0.9, weight decay set to 2e-4, and initial learning rate set to 1e-6 which is divided by 10 after 23,000 iterations.
Our network is trained for 30,000 iterations.
We evaluate our GearNet on the Berkeley Segmentation Dataset and Benchmark (BSDS 500) \cite{arbelaez2011contour}, which is one of notable benchmarks in the edge detection field.
This dataset contains 200 training, 100 validation, and 200 testing images, each with accurately
annotated boundaries.
Besides, our training set also incorporates the images from the PASCAL Context Dataset \cite{mottaghi2014role} and performs data augmentation
as in \cite{liu2016richer,xie2015holistically} for fair comparisons.
Similar to previous works, we use the fixed contour threshold (ODS) and per-image best threshold (OIS)
as our measures.
Before evaluation, we apply the standard non-maximal suppression algorithm to get thinned edges.

\subsection{Ablation Analysis}

\myPara{The Number of Encoders.}
The number of encoders also plays an important role in the edge detection task.
We consider the RCF network \cite{liu2016richer} as a special case of GearNet with 0 encoders.
From Table~\ref{tab:edge_comps}, we observe that adding 1 encoder (the bottom row) based on
the RCF architecture helps us obtain an increase of 0.4\% in terms of ODS.
When we add 2 middle encoders, the ODS score can be further improved by 0.3 points.
We observed no significant improvement when adding more than two encoders.

\renewcommand{\addFig}[1]{\includegraphics[width=0.163\linewidth]{#1}}

\begin{figure*}[t]
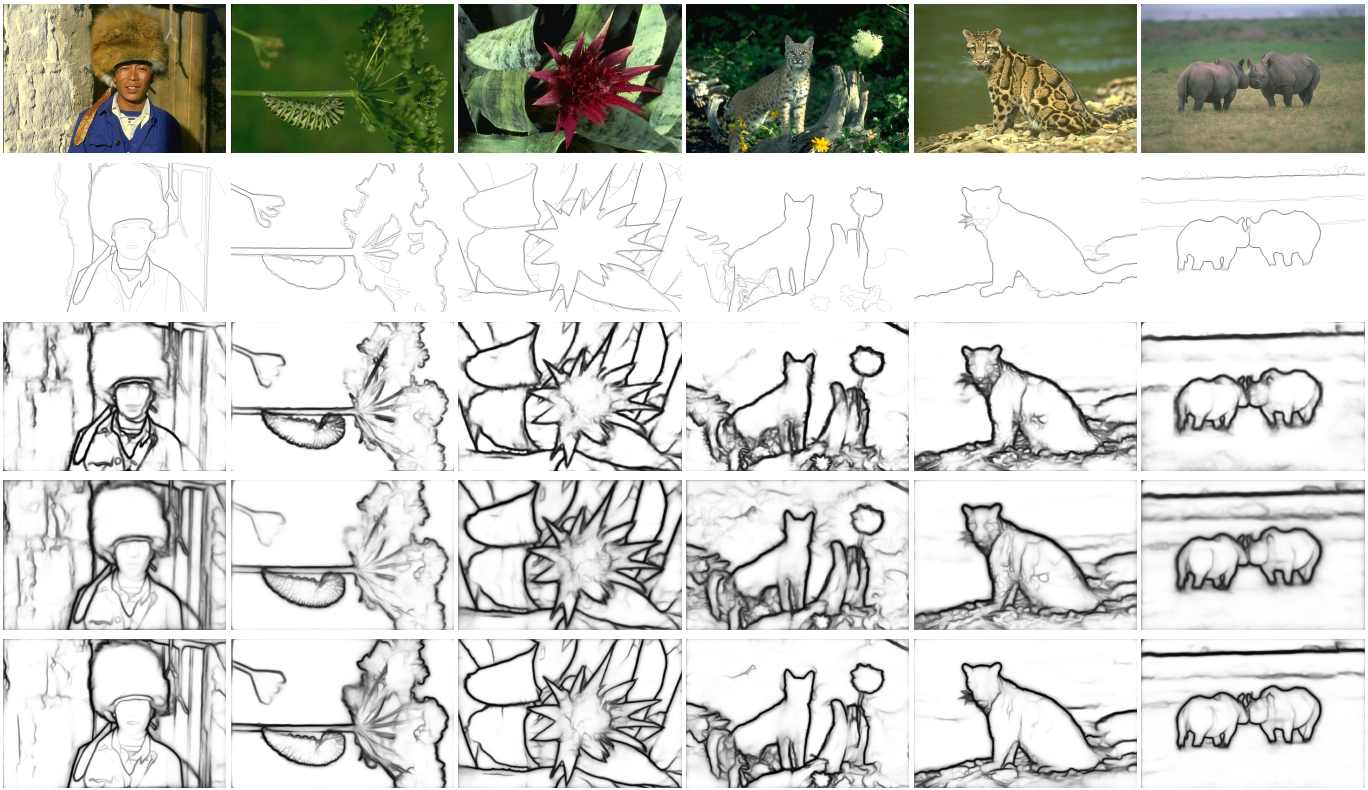

     \centering
     %\scriptsize
     \renewcommand{\tabcolsep}{1pt}
     \begin{tabular*}{1.0\textwidth}{cccccc}
        % \begin{sideways} ~~~~~~~Source \end{sideways} &
        \addFig{edge_samples/15062.jpg} &
        \addFig{edge_samples/35028.jpg} &
        \addFig{edge_samples/51084.jpg} &
        \addFig{edge_samples/326025.jpg} &
        \addFig{edge_samples/160067.jpg} &
        \addFig{edge_samples/112056.jpg} \\

        \addFig{edge_samples/15062_gt.png} &
        \addFig{edge_samples/35028_gt.png} &
        \addFig{edge_samples/51084_gt.png} &
        \addFig{edge_samples/326025_gt.png} &
        \addFig{edge_samples/160067_gt.png} &
        \addFig{edge_samples/112056_gt.png} \\

        \addFig{edge_samples/15062_hed.png} &
        \addFig{edge_samples/35028_hed.png} &
        \addFig{edge_samples/51084_hed.png} &
        \addFig{edge_samples/326025_hed.png} &
        \addFig{edge_samples/160067_hed.png} &
        \addFig{edge_samples/112056_hed.png} \\

        \addFig{edge_samples/15062_rcf.png} &
        \addFig{edge_samples/35028_rcf.png} &
        \addFig{edge_samples/51084_rcf.png} &
        \addFig{edge_samples/326025_rcf.png} &
        \addFig{edge_samples/160067_rcf.png} &
        \addFig{edge_samples/112056_rcf.png} \\

        \addFig{edge_samples/15062_mtp.png} &
        \addFig{edge_samples/35028_mtp.png} &
        \addFig{edge_samples/51084_mtp.png} &
        \addFig{edge_samples/326025_mtp.png} &
        \addFig{edge_samples/160067_mtp.png} &
        \addFig{edge_samples/112056_mtp.png} \\

        % (a) Image & (b) GT & (c) HED \cite{xie2015holistically} & (d) RCF \cite{liu2016richer} & (e) Ours
     \end{tabular*}
     \caption{Visual comparisons with several recent state-of-the-art edge detectors. From top to bottom:
     the source images; human-annotated ground truths; results by HED \cite{xie2015holistically};
     results by RCF \cite{liu2016richer}; our results. As can be seen, our proposed approach is able to
     not only generate cleaner background but also capture weak object boundaries compared to the other
     two methods. This phenomenon is specially clear for the second image.
     All the images are from the BSDS 500 dataset \cite{arbelaez2011contour}.}
     \label{fig:edge_vis}
     %\vspace{-10pt}
 \end{figure*}

\myPara{The Number of Channels.}
As stated in Sec.~\ref{sec:impl_details}, the convolutional layers in each side path are all with 16 channels.
To explore how the channel numbers used in each side path effect the performance of our architecture,
we also attempt to increase the channel numbers as done in \cite{liu2016richer} (21 channels).
However, the results show that more channels in each side path gives worse performance, leading to
a decrease of around 0.2 points in terms of ODS.
Similar phenomenon was also encountered when decreasing the number of channels.

 \begin{table}[!t]
    \centering
    \renewcommand{\tabcolsep}{1.2mm}
    \renewcommand{\arraystretch}{1.3}
    \caption{The results when different horizontal signal flow patterns are used for edge detection.
    We report both ODS score and OIS score. As can be observed, edge detection is more sensitive to
    the number of horizontal signal flows.}
    \label{tab:edge_ablation}
    \begin{tabular}{ccccccc} \toprule[1pt]
        %Tasks & Ours & Other Methods \\ \hline
        & \multicolumn{2}{c}{Pattern 1} & \multicolumn{2}{c}{Pattern 2} & \multicolumn{2}{c}{Pattern 3} \\ \cmidrule(l){2-3}\cmidrule(l){4-5}\cmidrule(l){6-7}%\multicolumn{2}{c}{stride} \\
        %\cmidrule(l){1-2} \cmidrule(l){3-4} \cmidrule(l){5-6}
        bottom & $S_1$ & $S_2$ & $S_1$ & $S_2$ & $S_1$ & $S_2$  \\ \midrule[1pt]
        conv1 & $T_1^{\{1, 2, 3\}}$ & $T_2^{\{1, 2\}}$ & $T_1^{\{1, 2, 3\}}$ & $T_2^{\{1\}}$ & $T_1^{\{1, 2, 3, 4\}}$ & $T_2^{\{1, 2, 3, 4\}}$ \\
        conv2 & $T_1^{\{2, 3, 4\}}$ & $T_2^{\{2, 3\}}$ & $T_1^{\{2, 3, 4\}}$ & $T_2^{\{2\}}$ & $T_1^{\{2, 3, 4, 5\}}$ & $T_2^{\{2, 3, 4\}}$    \\
        conv3 & $T_1^{\{3, 4, 5\}}$ & $T_2^{\{3, 4\}}$ & $T_1^{\{3, 4, 5\}}$ & $T_2^{\{3\}}$ & $T_1^{\{3, 4, 5\}}$    & $T_2^{\{3, 4\}}$    \\
        conv4 & $T_1^{\{4, 5\}}$    & -                & $T_1^{\{4, 5\}}$    & $T_2^{\{4\}}$ & $T_1^{\{4, 5\}}$       & -                   \\
        conv5 & - & - & - & - & - & -  \\ \midrule[1pt]
        ODS & \multicolumn{2}{c}{0.814} & \multicolumn{2}{c}{0.812} & \multicolumn{2}{c}{0.815} \\
        OIS & \multicolumn{2}{c}{0.835} & \multicolumn{2}{c}{0.832} & \multicolumn{2}{c}{0.834} \\ \bottomrule[1pt]
    \end{tabular}
\end{table}

\myPara{The Effect of Horizontal Signal Flows.}
%Besides the default structure,
We also analyze the number of horizontal signal flows in this paragraph.
While more horizontal signal flows helps salient object segmentation,
we found that this operation does not boost the performance in edge detection.
In Table~\ref{tab:edge_ablation}, we attempt to simplify our network by 
reducing the inputs of each transition node (Patterns 1 and 2).
According to the experimental results, these two patterns degrade the performance by 
nearly 0.4 and 0.6 points in ODS, respectively.
Furthermore, when we try to increase the number of horizontal signal flows in our architecture
as in Pattern 3 in Table~\ref{tab:edge_ablation}, the performance decreases as well.
This demonstrates different kinds of low-level features are essential to edge detection.
However, too many high-level features also harm the quality of the predicted edges because of
the lack of rich detailed information.
%
% More experimental settings and results can be found in our supplementary
% material.
%When we adopt a similar structure to the edge part in Table~\ref{tab:hh_info},
%the performance drops a little (0.5 points) in F-measure.
%
%When we reduce more horizontal signal flows, the performance decreases further.
%
%This phenomenon indicates that more connections between transition nodes do help for salient object detection
%such segment detection type tasks.

%\myPara{Data Augmentation.}
%Data augmentation as pointed out in previous works \cite{xie2015holistically} is useful for edge detection.
%%
%As a result, other than training on the training set of BSDS 500, we also add the images from the PASCAL Context
%Dataset \cite{mottaghi2014role} for training as in \cite{liu2016richer}.

\subsection{Comparison with the State-of-the-Art}

We compare our results with results from 12 existing methods, including gPb-owt-ucm \cite{arbelaez2011contour}, SE-Var \cite{dollar2015fast}, MCG \cite{pont2017multiscale}, DeepEdge \cite{bertasius2015deepedge}, DeepContour \cite{shen2015deepcontour}, HED \cite{xie2015holistically}, CEDN \cite{yang2016object}, RDS \cite{liu2016learning}, COB \cite{maninis2017convolutional}, CED \cite{wang2017deep}, DCNN+sPb \cite{kokkinos2015pushing}, and RCF-MS \cite{liu2016richer}, most of which are CNN-based methods.

 \begin{figure}[t]
     \centering
     \footnotesize
     \renewcommand{\tabcolsep}{2mm}
     \begin{tabular}{c}
        \includegraphics[width=0.9\linewidth]{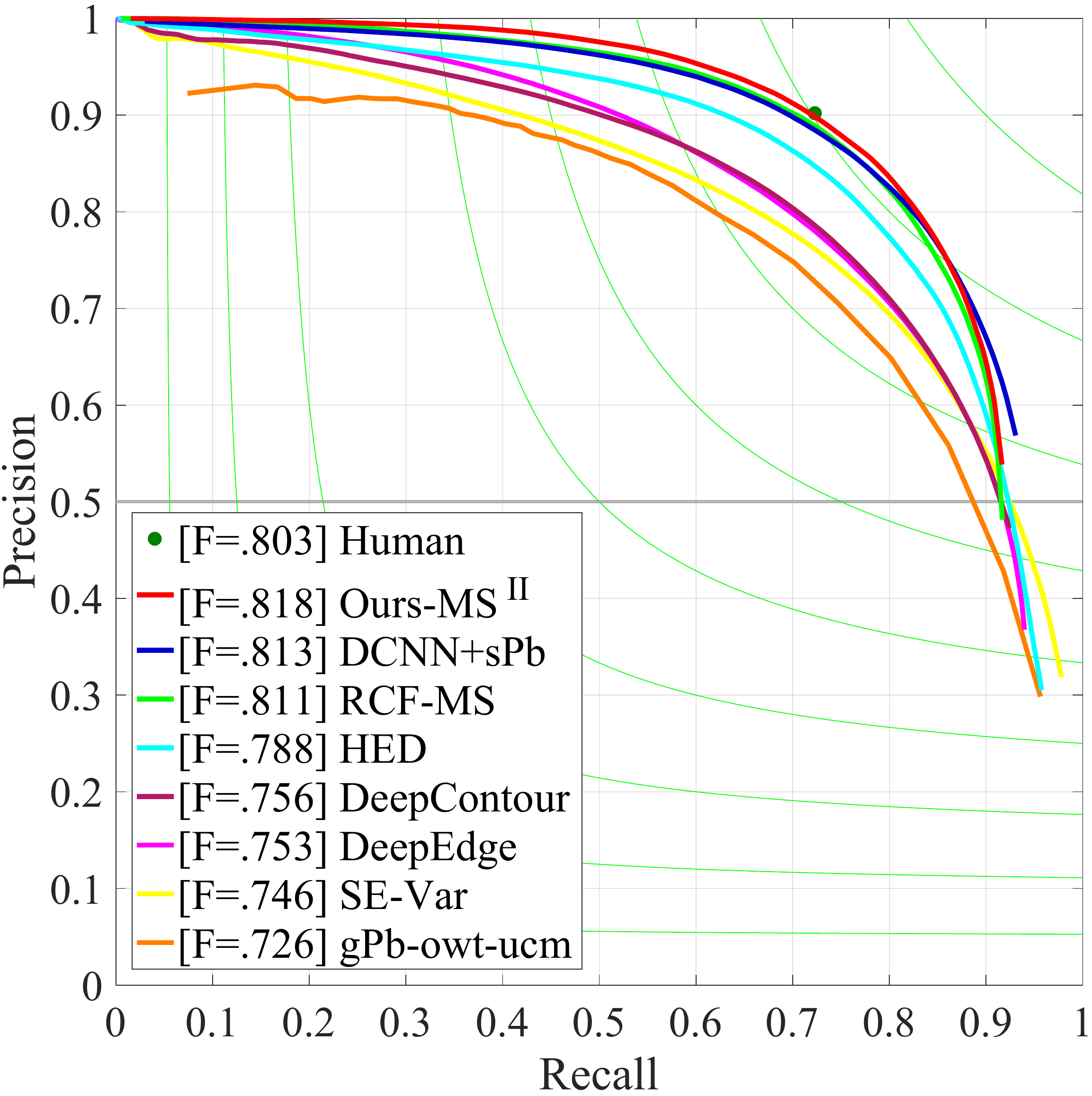}
     \end{tabular}
     \vspace{2pt}
     \caption{The precision-recall curves on BSDS 500 dataset \cite{arbelaez2011contour}.}
     \label{fig:edge_pr}
 \end{figure}

\myPara{Quantitative Analysis.} In Table~\ref{tab:edge_comps}, we show the quantitative results
of 12 previous works as well as ours.
With only one encoder, our method achieves ODS of 0.815 and OIS of 0.834, which are
already better than the most of the previous works.
This indicates that fusing features from different blocks of VGGNet performs better than
combining only the feature maps from the same block \cite{liu2016learning}.
On the other hand, more side supervision does help learning rich feature representations
\cite{maninis2017convolutional}.
Furthermore, when we add another encoder to our architecture as in Table~\ref{tab:hh_info}, our results
can be further enhanced from ODS of 0.815 to ODS of 0.818 $(+0.003)$.
For OIS, a similar phenomenon can also be found in Table~\ref{tab:edge_comps}.

\myPara{Visual Analysis.} In \figref{fig:edge_vis}, we show some visual comparisons between our approach and a leading representative method \cite{liu2016learning,xie2015holistically}.
As can be observed, our approach performs better in detecting the boundaries compared to the other one.
In \figref{fig:edge_vis}b, it is apparent that the real boundaries of the plants are highlighted well.
In addition, thanks to the fusion mechanism in our approach, the features learned by our network
are much more powerful compared to \cite{xie2015holistically,liu2016learning}.
This is because the areas with no edges are rendered much cleaner, especially in Figs.~\ref{fig:edge_vis}a and
\ref{fig:edge_vis}b.
To sum up, in spite of less than 1 point improvement compared to \cite{liu2016learning},
the quality of our results is much higher visually.

\myPara{PR Curve Comparisons.}
 The precision-recall curves of some selected methods can be found in \figref{fig:edge_pr}.
 One can observe that the PR-Curve produced by our approach is already better than
 humans in some certain cases and is better than all previous methods.

\begin{table}[tp]
    \centering
    %\scriptsize
    \renewcommand{\arraystretch}{1}
    \caption{Quantitative comparisons with existing skeleton extraction methods.
        The best results are highlighted in \textbf{bold}.}
    \label{tab:sk_comps}
    \begin{tabular}{L{28mm}C{26mm}C{26mm}} \toprule[1pt]%\whline{1pt}%\cline{1-2}
      & \multicolumn{2}{c}{Skeleton Datasets (F-measure)}  \\ \cmidrule(l){2-3}
      Method & SK-LARGE & WH-SYMMAX \\ \midrule[1pt]
        %MIL \cite{tsogkas2012learning} & 29.3\% & 17.4\% \\ \midrule[-0.7mm]
        HED \cite{xie2015holistically} & 49.7\% & 73.2\% \\ \midrule[-0.7mm]
        FSDS \cite{shen2016object} & 63.3\% & 76.9\% \\ \midrule[-0.7mm]
        LMSDS \cite{shen2017deepskeleton} & 64.9\% & 77.9\% \\ \midrule[-0.7mm]
        SRN \cite{ke2017srn} & 61.5\% & 78.0\% \\ \midrule
        %----------------------------------------------------------------
        %$\text{GearNet}^{\mathrm{II}*}$ (Ours) & 60.2\% & 61.4\% & - \\ \midrule[-0.7mm]
        %$\text{GearNet}^{\mathrm{II}}$ (Ours) & - & - \\ \midrule[-0.7mm]
        $\text{GearNet}$ (Ours) & \textbf{68.3\%} & \textbf{80.1\%} \\ \midrule[-0.7mm]
        %$\text{GearNet-MS}$ (Ours) & \textbf{71.0\%} & - \\ %\midrule[-0.7mm]
        \bottomrule[1pt]
    \end{tabular}
\end{table}

\section{Application III: Skeleton Extraction}\label{sec:skeleton}

In this section, we apply our GearNet to skeleton extraction.
%
%Different from salient object detection and edge detection, skeleton extraction is an unknown-scale
%problem in that the thickness of skeletons is associated with the corresponding object parts.
%
We will show that our method substantially outperforms priors works by a large margin.

\subsection{Ablation Analysis}
The hyper-parameters we use are as follows: weight decay set to 0.0002, momentum set to 0.9, mini-batch size set to 10, and initial learning rate of 1e-6 which is divided by 10 after 20,000 iterations.
We run the network for 30,000 iterations.
A standard non-maximal suppression algorithm is used to obtain thinned skeletons before evaluation.
We use the same training set to \cite{shen2017deepskeleton}. 

\myPara{The Number of Encoders.} As in salient object detection and edge detection, reducing the number
of encoders when carrying out skeleton detection degrades the performance.
When we remove the second middle encoder $\mathcal{E}_2$, the F-measure score drops by more than 2 points.
Adding more encoders in our approach also leads to no performance gain.

\myPara{The Number of Channels.}
In our experiment, similarly to edge detection, we also try to reduce the number of channels 
in each side path to half of each and observe that the results slightly decrease
by 0.5 points.
When we further reduce the channel numbers to a quarter of each, the performance drops dramatically
(by more than 5 points).
This phenomenon indicates that skeleton extraction relies on more information from the backbone.
This can be achieved by adding proper number of channels in each side path.
In addition, we also attempt to increase the number of channels by doubling them but 
find no performance gain.

\begin{figure}[t]
     \centering
     \scriptsize
     \renewcommand{\tabcolsep}{0.3mm}
     % \scalebox{0.8}{
     \begin{tabular*}{1.0\textwidth}{cccc}
        %\begin{sideways} ~~FSDS \cite{shen2016object} \end{sideways} &
        \includegraphics[width=0.245\linewidth]{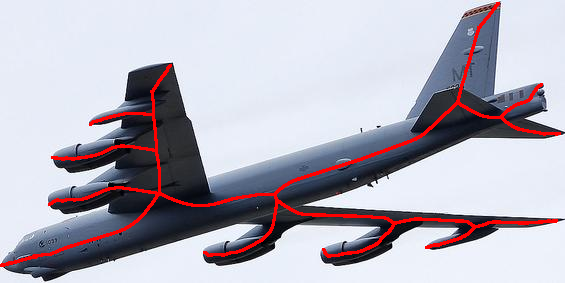} &
        \includegraphics[width=0.245\linewidth]{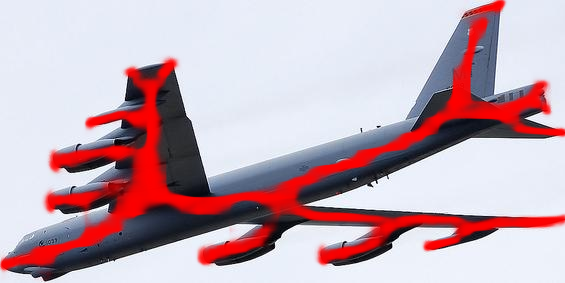} &
        \includegraphics[width=0.245\linewidth]{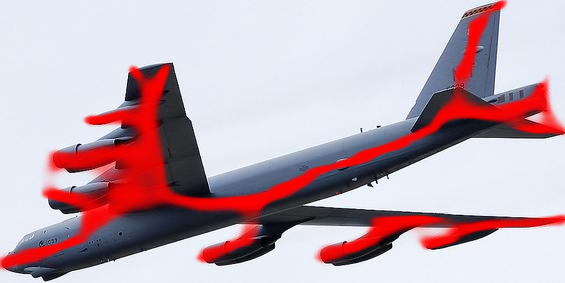} &
        \includegraphics[width=0.245\linewidth]{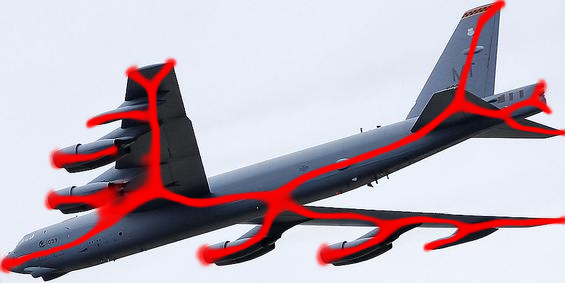} \\
        %\begin{sideways} ~~~~~SRN \cite{ke2017srn} \end{sideways} &
        \includegraphics[width=0.245\linewidth]{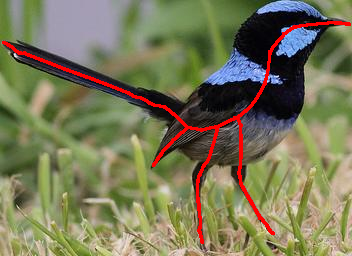} &
        \includegraphics[width=0.245\linewidth]{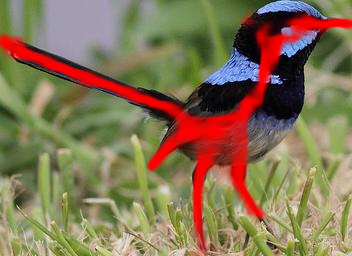} &
        \includegraphics[width=0.245\linewidth]{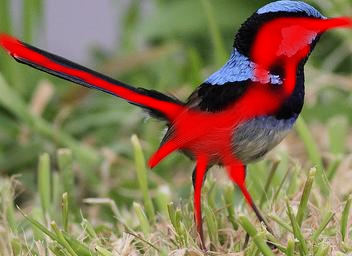} &
        \includegraphics[width=0.245\linewidth]{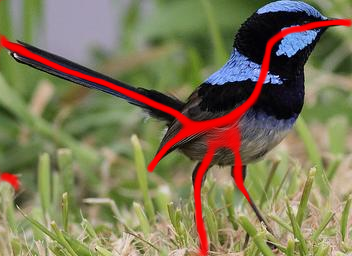} \\
        %\begin{sideways} ~~~~Ours \end{sideways} &
        \includegraphics[width=0.245\linewidth]{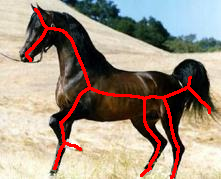} &
        \includegraphics[width=0.245\linewidth]{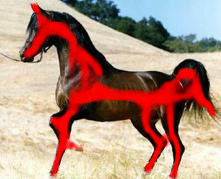} &
        \includegraphics[width=0.245\linewidth]{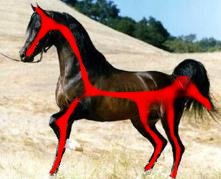} &
        \includegraphics[width=0.245\linewidth]{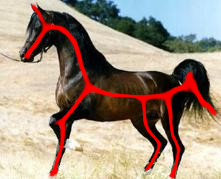} \\

        \includegraphics[width=0.245\linewidth]{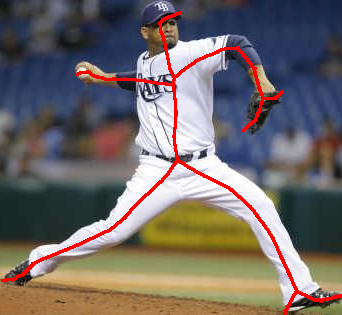} &
        \includegraphics[width=0.245\linewidth]{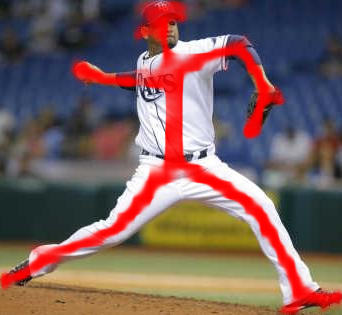} &
        \includegraphics[width=0.245\linewidth]{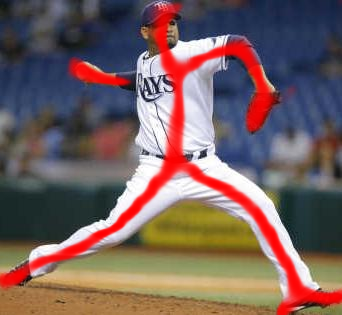} &
        \includegraphics[width=0.245\linewidth]{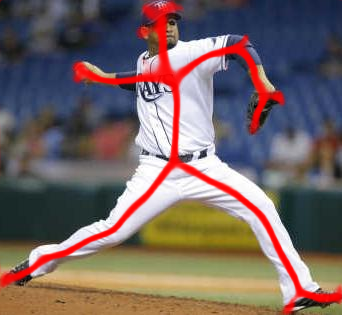} \\

        \includegraphics[width=0.245\linewidth]{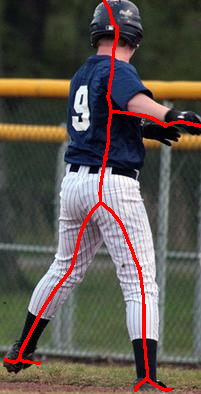} &
        \includegraphics[width=0.245\linewidth]{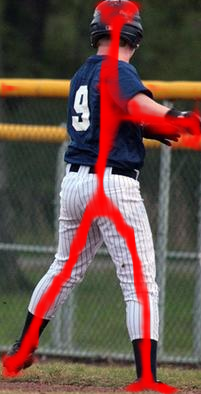} &
        \includegraphics[width=0.245\linewidth]{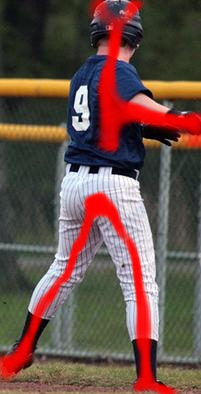} &
        \includegraphics[width=0.245\linewidth]{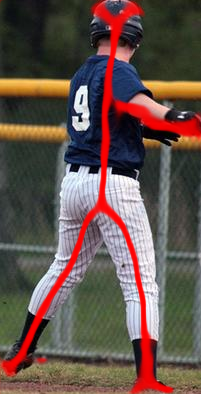} \\
        (a) GT & (b) FSDS \cite{shen2016object} & (c) SRN \cite{ke2017srn} & (d) Ours
     \end{tabular*}
     % }
     \caption{Visual comparisons with two recently representative skeleton extraction approaches.
     It can be easily found that our results are much thinner than the other two methods. Also,
     the skeletons produced by our results are continuous, which is essential for its applications.}
     \label{fig:sk_vis}
     \vspace{-10pt}
 \end{figure}

\myPara{The Effect of Horizontal Signal Flows.}
The number of horizontal signal flows also affects the skeleton results.
We reduce the number of optional connections between $\mathcal{E}_1$ and $\mathcal{E}_2$ 
from 3 to 2 but the F-measure score decreases by around 2 points.
Reducing the number of optional connections between $\mathcal{E}_0$ and
$\mathcal{E}_1$ leads to a similar phenomenon.
This indicates introducing connections between higher and lower layers is essential
for skeleton detection.

 \begin{figure*}[t]
     \centering
     \footnotesize
     \renewcommand{\tabcolsep}{2mm}
     \begin{tabular}{c}
        \includegraphics[width=0.45\linewidth]{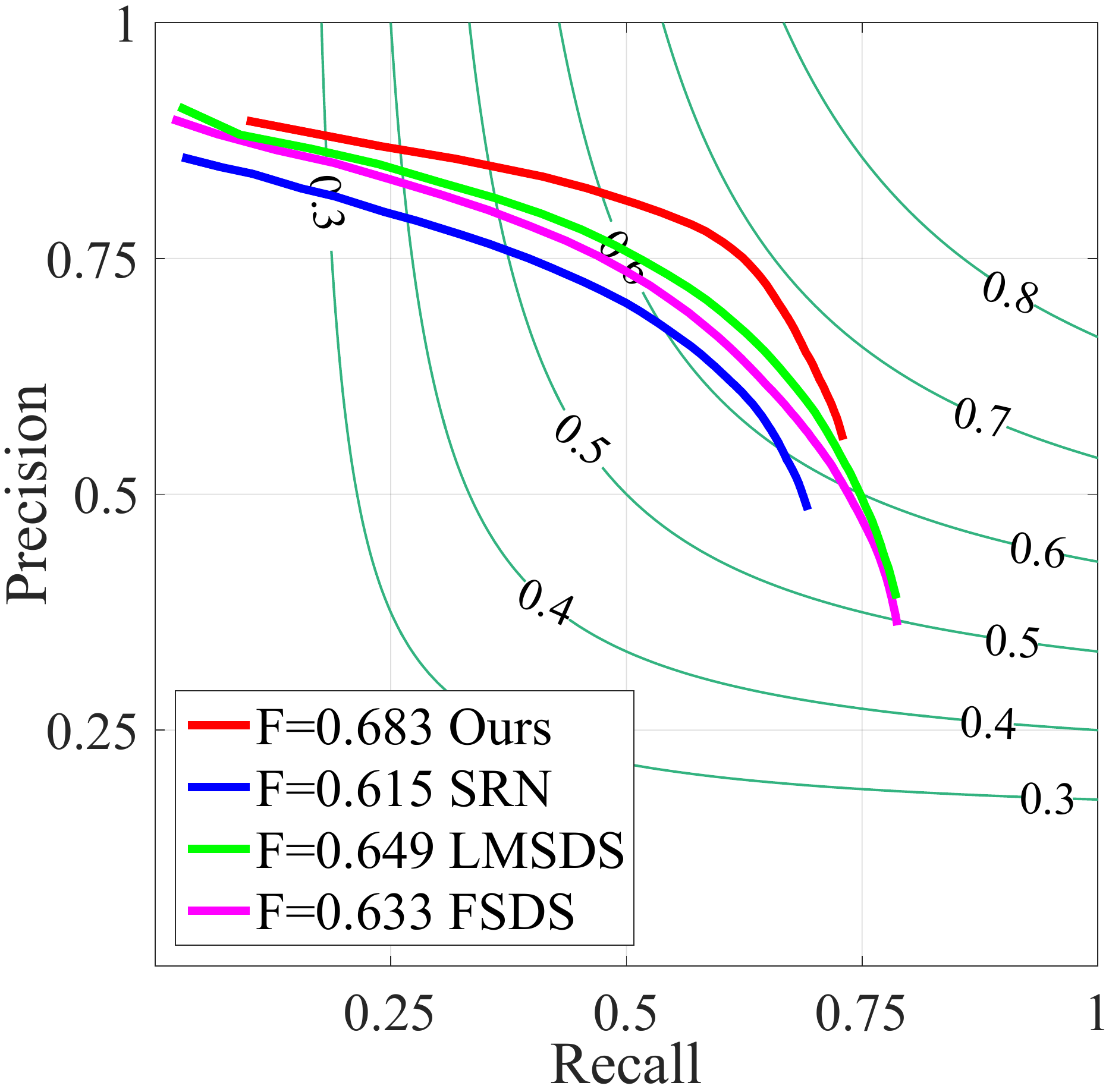} \quad\quad
        \includegraphics[width=0.45\linewidth]{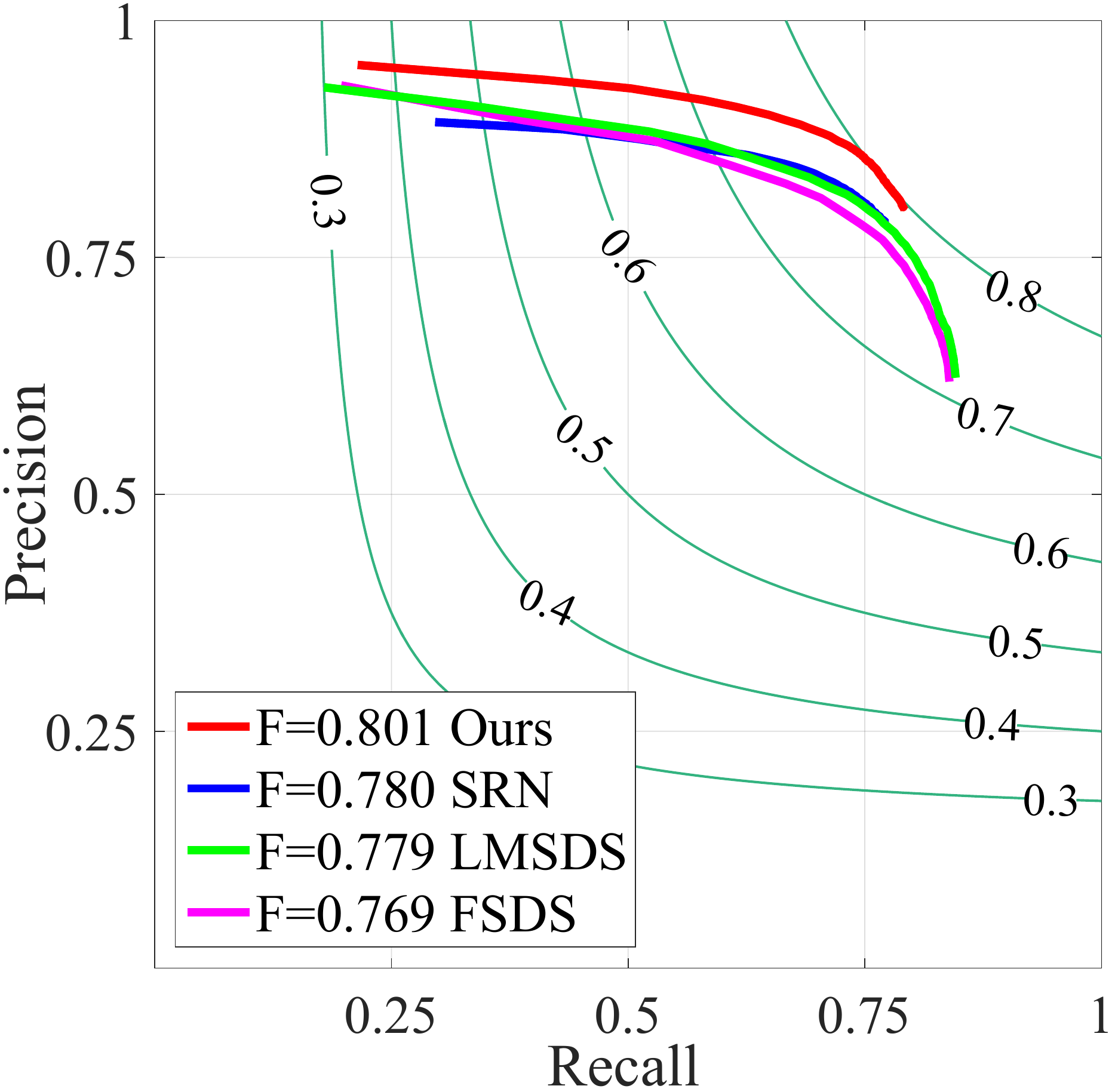}
     \end{tabular}
     \caption{The precision-recall curves on SK-LARGE dataset \cite{shen2017deepskeleton}
     and WH-SYMMAX dataset \cite{shen2016multiple}. It can be easily observed that our approach
     greatly outperforms existing state-of-the-art methods.}
     \label{fig:skel_sklarge_pr}
 \end{figure*}

\subsection{Comparison with the State-of-the-Arts}
We compare our GearNet with 4 recent CNN-based methods (HED \cite{xie2015holistically}, FSDS \cite{shen2016object}, LMSDS \cite{shen2017deepskeleton}, and SRN \cite{ke2017srn}) on 2 popular and challenging datasets including SK-LARGE \cite{shen2017deepskeleton} and WH-SYMMAX \cite{shen2016multiple}.
Similar to \cite{shen2016object}, we use the F-measure score to evaluate the quality of prediction maps.
In Table~\ref{tab:sk_comps}, we show quantitative comparisons with existing methods.
As can be seen, our method wins dramatically by a large margin (3.4 points) on the SK-LARGE dataset \cite{shen2016object}.
There is also an improvement of 2.1 points on the WH-SYMMAX dataset \cite{shen2016multiple}.
%
%When multi-scale testing is considered, another 2 points improvement is obtained.
%
%
In \figref{fig:sk_vis}, we also show some visual illustrations of three approaches (more can be found in
our supplementary materials).
Owing to the advanced features extracted from our GearNet, our method is able to more accurately locate
the exact positions of the skeletons.
This point can also be substantiated by the fact that our prediction maps are also much thinner than other works.
Both quantitative and visual results unveil that our horizontal hierarchy provides a better way
to combine different-level features for skeleton extraction.
%
%We also believe that its flexible structure allows it to be applied to a wide variety of binary tasks
%in computer vision.

\myPara{PR Curve Comparisons}
In \figref{fig:skel_sklarge_pr}, we also show the precision-recall curses on two datasets, including SK-LARGE \cite{shen2017deepskeleton} and WH-SYMMAX \cite{shen2016multiple}.
As can be seen, quantitatively, our approach on both datasets substantially outperforms other existing methods, which verify the statement we explain above.

%---------------------------------------------------------------

\section{Discussion and Conclusion}\label{sec:discussion}
%\vspace{-5pt}
In this paper, we present a novel architecture for binary vision tasks and apply it to
three drastically different example tasks including salient object segmentation, edge detection, and skeleton extraction.
Notice, however, that our approach is not limited to these tasks and can be potentially applied to a wide variety of binary pixel labeling tasks in computer vision.
In order to take more advantage of CNNs, we introduce the concept of transition node, which receives signals from different-level features maps.
%%
%By adding several stages of transition nodes, our network is able to build rich feature
%representations.
%%
%%We apply our proposed unified framework to two fundamental binary problems---salient object
%%detection and edge detection.
%%
Exhaustive evaluations and comparisons with recent notable state of the art methods on widely used datasets shows that our framework outperforms all of them in all three tasks, testifying the power of our proposed framework.
Further, we structurally analyze our proposed architecture using several ablation experiments in each task and investigate the roles of different design choices in our approach. We hope that our work will encourage subsequent research to design universal architectures for computer vision tasks.

% or


% \appendices

% % you can choose not to have a title for an appendix
% % if you want by leaving the argument blank
% \section{}

\section*{Acknowledgments}
This research was supported by the National Youth Talent Support Program,
NSFC (NO. 61620106008, 61572264),
Tianjin Science Fund for Distinguished Young Scholars (17JCJQJC43700),
and Huawei Innovation Research Program.

\ifCLASSOPTIONcaptionsoff
  \newpage
\fi



%\bibliographystyle{IEEEtran}
%\begin{thebibliography}{1}


{
\bibliographystyle{IEEEtran}
\bibliography{GearNet}
}

\vspace{-.3in}
\begin{IEEEbiography}[{\includegraphics[width=1in,keepaspectratio]{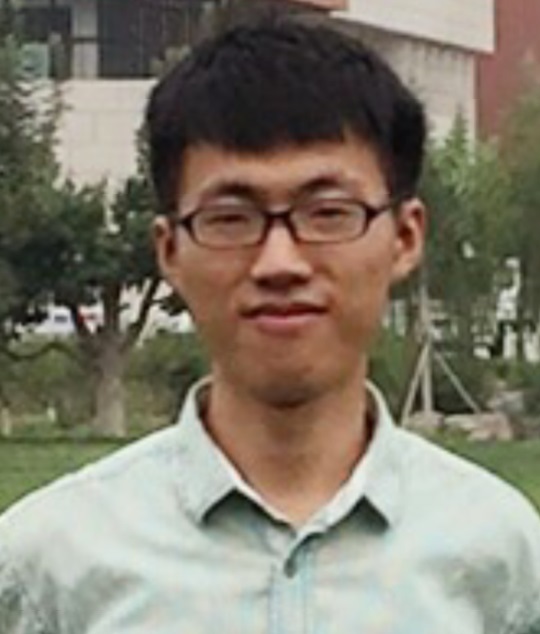}}]
{Qibin Hou} is currently a Ph.D. Candidate with College of Computer Science and
Control Engineering, Nankai University, under the supervision of Prof. Ming-Ming Cheng.
His research interests include deep learning, image processing, and computer vision.
\end{IEEEbiography}

\vspace{-.5in}
\begin{IEEEbiography}[{\includegraphics[width=1in,keepaspectratio]{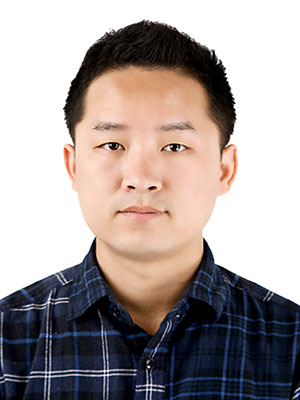}}]
{Ming-Ming Cheng} received his PhD degree from Tsinghua University in 2012.
Then he did 2 years research fellow, with Prof. Philip Torr in Oxford.
He is now a professor at Nankai University, leading the Media Computing Lab.
His research interests includes computer graphics, computer vision, and image processing.
He received research awards including ACM China Rising Star Award,
IBM Global SUR Award, CCF-Intel Young Faculty Researcher Program, etc.
\end{IEEEbiography}

\vspace{-.4in}
\begin{IEEEbiography}[{\includegraphics[width=1in,keepaspectratio]{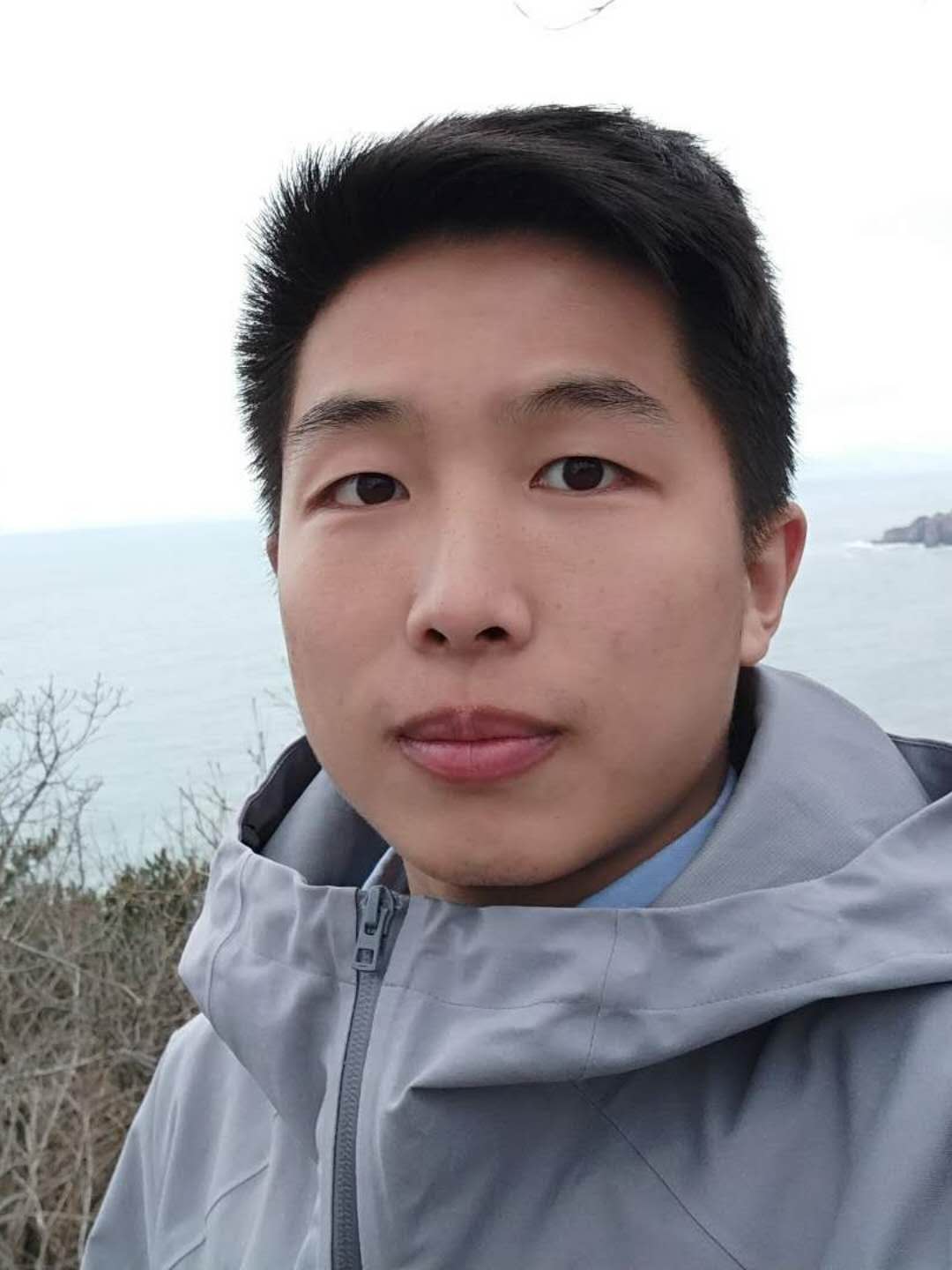}}]
{Jiangjiang Liu} is currently a Master student
with College of Computer Science and
Control Engineering, Nankai University,
under the supervision of Prof. Ming-Ming Cheng.
His research interests include deep learning,
image processing, and computer vision.
\end{IEEEbiography}

\vspace{-.4in}
\begin{IEEEbiography}[{\includegraphics[width=1in,keepaspectratio]{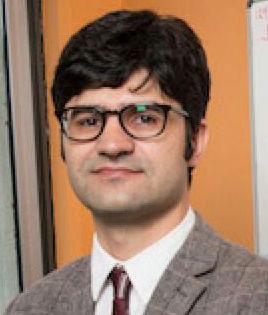}}]
{Ali Borji}
received the PhD degree in cognitive
neurosciences from the Institute for Studies in
Fundamental Sciences (IPM), 2009.
He is currently an assistant professor at
Center for Research in Computer Vision,
University of Central Florida.
His research interests include visual
attention, visual search, machine learning, neurosciences,
and biologically plausible vision models.
\end{IEEEbiography}

\vspace{-.4in}
\begin{IEEEbiography}[{\includegraphics[width=1in,keepaspectratio]{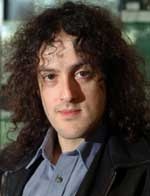}}]
{Philip H.S. Torr} received the PhD degree from Oxford University.
After working for another 3 years at Oxford,
he worked for 6 years as a research scientist for Microsoft Research,
first in Redmond, then in Cambridge,
founding the vision side of the Machine Learning and Perception Group.
He is now a professor at Oxford University.
He has won awards from several top vision conferences,
including ICCV, CVPR, ECCV, NIPS, etc.
He is a Royal Society Wolfson Research Merit Award holder.
\end{IEEEbiography}

\end{document}